\newif\ifpreprintversion
 \preprintversiontrue %

\documentclass[conference]{IEEEtran}

\ifCLASSINFOpdf
\else
\fi

\usepackage{cite}
\usepackage[numbers, square]{natbib}
\usepackage[hidelinks]{hyperref}

\usepackage[show]{chato-notes}
\usepackage{subfig}
\usepackage{graphicx} %
\usepackage{svg}
\usepackage{booktabs}
\usepackage{amsmath}
\usepackage{amsfonts}  %
\usepackage{url}
\usepackage[capitalize,noabbrev]{cleveref}
\usepackage{multirow}
\usepackage[ruled]{algorithm2e}
\usepackage[]{mdframed}
\usepackage{tabularx}

\usepackage{tcolorbox}%
\definecolor{framecolor}{rgb}{0.122, 0.435, 0.698}%
\definecolor{bgcolor}{rgb}{0.95, 0.95, 0.95}%

\newcommand{\goalbox}[1]{%
  \begin{tcolorbox}[
    colframe=framecolor,       %
    colback=bgcolor,           %
    boxrule=0pt,               %
    leftrule=3pt,              %
    arc=0pt,                   %
    left=2pt,                  %
    right=2pt,                 %
    top=2pt,                   %
    bottom=2pt,                %
    width=\linewidth,          %
    before skip=4pt,           %
    after skip=4pt             %
  ]
    #1
  \end{tcolorbox}
}

\newcommand{\eg}{e.g.,\xspace}
\newcommand{\ie}{i.e.,\xspace}

\newcommand\shortsection[1]{\vspace{6pt}{\noindent\textbf{#1.}}}

\newcommand{\revision}[1]{\textcolor{black}{#1}}

\begin{document}

\title{DROP: Poison Dilution via Knowledge Distillation for Federated Learning}

\ifpreprintversion
\author{
\IEEEauthorblockN{Georgios Syros$^{*\dagger1}$, Anshuman Suri$^{*1}$, Farinaz Koushanfar$^{2}$, Cristina Nita-Rotaru$^{1}$, Alina Oprea$^{1}$}
	\IEEEauthorblockA{Northeastern University$^{1}$, University of California, San Diego$^{2}$}

}
\else
\author{
    \IEEEauthorblockN{Anonymous Submission}
}

\fi

\ifpreprintversion

\else

\IEEEoverridecommandlockouts
\makeatletter\def\@IEEEpubidpullup{6.5\baselineskip}\makeatother
\IEEEpubid{\parbox{\columnwidth}{
		Network and Distributed System Security (NDSS) Symposium 2025\\
		24-28 February 2025, San Diego, CA, USA\\
		ISBN 979-8-9894372-8-3\\
		https://dx.doi.org/10.14722/ndss.2025.[23$|$24]xxxx\\
		www.ndss-symposium.org
}
\hspace{\columnsep}\makebox[\columnwidth]{}}

\fi

\maketitle

\ifpreprintversion
    \renewcommand{\thefootnote}{}
    \footnotetext{$^*$ Equal Contribution}
    \footnotetext{$^\dagger$ Correspondence to syros.g@northeastern.edu}
    \newcommand{\codeurl}{\url{https://github.com/gsiros/drop}}
\else
    \newcommand{\codeurl}{\url{https://anonymous.4open.science/r/drop-035D/}}
\fi

\begin{abstract}
Federated Learning is vulnerable to adversarial manipulation, where malicious clients can inject poisoned updates to influence the global model's behavior. While existing defense mechanisms have made notable progress, they fail to protect against adversaries that aim to induce targeted backdoors under different learning and attack configurations. To address this limitation, we introduce \textit{DROP} (\textbf{D}istillation-based \textbf{R}eduction \textbf{O}f \textbf{P}oisoning), a novel defense mechanism that combines clustering and activity-tracking techniques with extraction of benign behavior from clients via knowledge-distillation to tackle stealthy adversaries that manipulate low data poisoning rates and diverse malicious client ratios within the federation. Through extensive experimentation, our approach demonstrates superior robustness compared to existing defenses across a wide range of learning configurations. Finally, we evaluate existing defenses and our method  under the challenging setting of non-IID client data distribution and highlight the challenges of designing a resilient FL defense in this setting.
\end{abstract}

\IEEEpeerreviewmaketitle

\section{Introduction}
\label{sec:introduction}

Machine learning (ML) systems greatly benefit from scaling the amount of data used for training ML models \cite{Covert2024ScalingLF, Parthasarathi2019RealizingPS, fl_benefits, Wu2021SustainableAE}. While centralized training is ideal, data contributors often hesitate to share their raw data with a centralized party. Federated learning (FL) \cite{kairouz2021advances} provides a middle-ground to help alleviate such privacy concerns, while potentially offloading compute onto clients. This design supposedly improves privacy, but it comes at the cost of less control over data and gradient ingestion, allowing malicious clients to introduce malicious goals into the global model through data poisoning \cite{gu_badnets, Wang2024LinkageOS} or gradient manipulation \cite{bagdasaryan_howto, Goldblum2020DatasetSF}.

This lack of control over client contributions enables a particularly insidious form of adversarial behavior known as \textit{backdoor poisoning attacks} \cite{gu_badnets, liu2018trojaning, Alam2023GetRO, Zhang2024ConcealingBM, Chai2023ASF}. Unlike traditional data poisoning, which aims to degrade overall model performance and mount an availability attack \cite{rosenfeld2020certified, biggio2013poisoningattackssupportvector, fangbyzantinerobust, Tolpegin2020DataPA}, backdoor attacks embed hidden triggers in the model that lead to selective misclassifications for certain inputs. This allows the global model to perform well on clean data while remaining vulnerable to targeted exploitation. The decentralized nature of FL amplifies this threat, as the server has limited visibility into local training data and model updates, making it difficult to distinguish between honest contributions and adversarial modifications. As a result, backdoor poisoning has emerged as one of the most concerning attack vectors in federated learning.

Although numerous defenses have been proposed to counter poisoning attempts in FL \cite{yin2018byzantine, blanchard2017machine, cao2021fltrust, fung2018mitigating, auror, wang2022flare, nguyen2022flame, zhang2023flip, mesas}, their evaluations lack consistency across different federation configurations and adversarial strategies. Many existing works focus on narrow or suboptimal configurations and attack scenarios, limiting their practical applicability \cite{khan2023pitfalls}. We identify multiple critical learning configurations---comprising variations in the local number of training epochs, learning rate, and batch size---where existing FL defenses fail to consistently mitigate targeted backdoor attacks. Small changes to these configurations can significantly influence the success of both the attack and the defense, underscoring the need for more robust and generalizable defensive solutions against backdoor poisoning attacks.

In this paper, we address the challenge of defending against \textit{targeted backdoor poisoning attacks}, a more sophisticated variant of traditional backdoor attacks where the adversary targets a specific ``victim'' class within the data distribution. This focused strategy makes the attack more stealthy and difficult to detect, as the backdoor trigger affects only a subset of inputs rather than the entire input space. To counter this threat, we propose \textbf{DROP}\footnote{Code available at \codeurl} (\textit{Distillation-based Reduction of Poisoning Signals}), a comprehensive, multi-layered defense framework. \revision{DROP introduces a novel architecture that combines effectively three key components}: (1) \textbf{Agglomerative Clustering} to identify and separate anomalous updates, (2) \textbf{Activity Monitoring} to track and penalize suspicious clients over multiple rounds, and (3) \textbf{Knowledge Distillation} to cleanse the global model using a synthetic dataset guided by the consensus logits from benign client updates. \revision{Through this novel set of countermeasures, DROP
provides resilience against both aggressive and stealthy poisoning attacks, across a wide range of  learning configurations.} Moreover, we propose DROPlet, a lightweight variant of DROP designed for seamless integration into existing FL defense frameworks. \revision{We evaluate DROP against seven existing defenses and three poisoning attacks across diverse FL configurations, demonstrating its robustness and adaptability.} Unlike existing defenses, which often rely on fixed client-selection strategies or specific attack assumptions, DROP generalizes to dynamic, real-world FL environments where adversarial strategies and learning configurations may vary significantly. Our results show that DROP effectively mitigates stealthy targeted backdoor attacks, reducing attack success rates across a broad spectrum of configurations.

\shortsection{Contributions} 
Our contributions are summarized as follows:  

\begin{itemize}  
    \item We evaluate seven existing defenses against targeted backdoor attacks and find that most defenses fail to provide adequate protection across a wide range of different learning configurations (learning rate, batch size, number of epochs), exhibiting high sensitivity to changes in the overall FL configuration (\Cref{sec:existing_finicky}).  
    
    \item Motivated by the lack of robust defenses against stealthy targeted backdoors, we propose DROP (\textit{Poison Dilution via Knowledge Distillation for Federated Learning}), a novel defense mechanism that leverages clustering and activity monitoring techniques with  knowledge distillation to tackle adversaries trying to exploit vulnerable learning configurations and different attack strategies (\Cref{sec:proposed_defense}).  
    \item Through extensive experimental evaluation across multiple \revision{backdoor attacks}, datasets and FL setups, we demonstrate that DROP offers enhanced protection against backdoor attacks from adversaries under varying degrees of stealthiness (\Cref{sec:experiments}). DROP shows resilience across a wide range of learning configurations and provides consistent attack mitigation throughout training rounds, outperforming existing defenses. For example, we observe an average attack success rate of just $1.93\%$ across 10 diverse FL configurations. However, defending against stealthy attacks under non-IID data distributions remains challenging, with most defenses struggling to offer complete protection in complex dataset scenarios.
\end{itemize}
Finally, we emphasize the need for researchers to assess robustness across diverse FL configurations (\Cref{sec:conclusion}).
We also release our source code as a comprehensive framework that includes implementations of existing defenses, enabling unified comparisons of attack and defense efficacy and facilitating the integration of new attacks and defenses.

\section{Background and Threat Model}
\label{sec:background}

We introduce the fundamentals of federated learning, describe the threat of poisoning attacks with a focus on backdoor attacks, and define the threat model considered in this work.

\subsection{Background}

\subsubsection{Federated Learning}

Federated Learning (FL) is a distributed machine learning paradigm where clients collaboratively train a global model while preserving the privacy of their local data, addressing regulatory and privacy concerns \citep{mcmahan2017communication}. The most widely used FL algorithm is Federated Averaging (\textbf{FedAvg}) \citep{mcmahan2017communication}, which operates iteratively through a series of communication rounds.  

At each round \(t\), the central server shares the global model parameters \(\mathbf{w}_t\) with a randomly selected subset of clients \(\mathcal{C}_t\). Each client \(c \in \mathcal{C}_t\) trains the model locally on its dataset \(\mathcal{D}_c\) and returns the updated parameters \(\mathbf{w}_c^t\). The server aggregates the updates to form the updated global model \(\mathbf{w}_{t+1}\):
\begin{equation}
    \mathbf{w}_{t+1} = \frac{1}{|\mathcal{C}_t|} \sum_{\substack{c \in \mathcal{C}_t}} \mathbf{w}_c^t.
\end{equation}
This process continues until the global model converges or a predefined number of rounds is completed.  

A typical FL system is parameterized by two categories of parameters: \textit{communication} parameters (\eg the number of clients \(C_t\) participating in each round \(t\)) and \textit{learning} parameters. Learning parameters, collectively referred to as the \textit{learning configuration}, define the local training process at each client and include the local \textit{learning rate}, the number of local training \textit{epochs}, and the \textit{batch size}. These parameters have a significant impact on the performance of the global model, which is typically measured using the \textit{Main Task Accuracy} (MTA)—the accuracy achieved by the global model on its intended learning task.

\subsubsection{Poisoning Attacks}

Poisoning attacks involve either degrading overall performance or introducing additional training objectives to  embed selective malicious behavior. In the context of FL, the lack of direct access to clients' raw data makes it harder to defend against such poisoning attacks. This opacity enables adversaries to inject malicious updates while remaining undetected.

Data poisoning, a common attack type, involves adversaries manipulating local training data to influence the global model. This includes injecting mislabeled or out-of-distribution examples to degrade performance (\textbf{untargeted attacks} \citep{biggio2013poisoningattackssupportvector, jagielski_manipulating, Mei_Zhu_2015, xiao_feature_selection, shejwalkar2021manipulating, fangbyzantinerobust}) or subverting predictions for specific tasks (\textbf{targeted attacks} \citep{koh_black_box, shafahi_poison_frogs, suciu_fail}). A more covert form, the \textbf{backdoor attack} \citep{gu_badnets, bagdasaryan_howto, attacktails, sun2019reallybackdoorfederatedlearning}, implants hidden behaviors, causing the model to misclassify inputs with a predefined trigger while maintaining accuracy on legitimate data.
\begin{figure}[h]
    \centering
    \includegraphics[width=0.9\columnwidth]{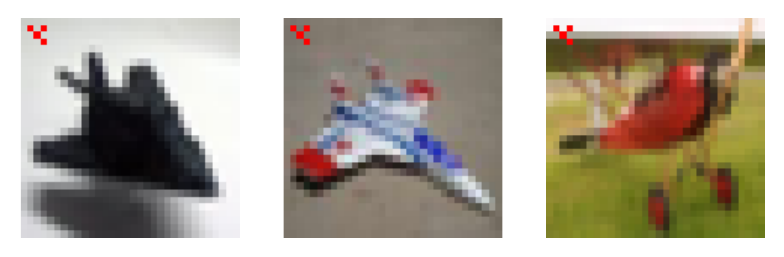}
    \caption{Examples of CIFAR-10 images from the \textit{plane} class with an added backdoor trigger (top-left corner). The presence of the trigger causes the model to misclassify these inputs as the \textit{horse} class, illustrating the effect of a targeted backdoor attack.}
    \label{fig:backdoor_samples}
\end{figure}

In this work, we focus on \textbf{targeted backdoor attacks} \citep{targeted_backdoors}, where adversaries poison a specific subpopulation (victim class) with a hidden \emph{trigger}. The backdoor activates only for victim class inputs containing the trigger, remaining covert while exploiting client update variability and the decentralized nature of FL systems, making detection challenging. An example is illustrated in \Cref{fig:backdoor_samples}.

\subsection{Threat Model}
\label{sec:threat_model}

\shortsection{Adversary's Goals} 
The adversary's primary objective is to embed a backdoor into the global model in a way that satisfies two key properties:
\newline
\newline
\underline{\textbf{Property 1}}: \textit{Stealthiness}. The backdoor should avoid arousing suspicion by not degrading performance on data without any triggers, whilst circumventing any defense measures the server may deploy.
\newline
\newline
\underline{\textbf{Property 2}}: \textit{Consistency}. The backdoor attack must maintain a consistently high \textit{Attack Success Rate} (ASR) when the model achieves acceptable MTA, ensuring the adversary's success is not reliant on opportunistic fluctuations in ASR across rounds. Specifically, let \(\text{MTA}_t\) and \(\text{ASR}_t\) denote the main task accuracy and attack success rate at round \(t\), respectively. We define the set of rounds \(K\) where the MTA is above some given threshold \(\lambda\) as:  
\begin{equation}
    K = \{t \mid \text{MTA}_t \geq \lambda\},
\end{equation}  
and require that the ASR across these rounds remains above a specified threshold \(\tau\):  
\begin{equation}
    \min_{t \in K} \text{ASR}_t \geq \tau.
\end{equation}  
This formulation ensures that the backdoor remains effective throughout training, preventing its success from being undermined by transient fluctuations in ASR.

\shortsection{Adversary's Capabilities} 
The adversary is successful in compromising and gaining control of a fraction $\rho$ of all clients \(\mathcal{N}\) in the system, which is referred to as the \textit{Malicious Client Ratio} (MCR). This set of compromised clients $\mathcal{N}_\text{adv}$ is fixed throughout the federation.
Once a client \(c \in \mathcal{N}_\text{adv}\) is compromised, the adversary gains full control over the client’s local training process, including its dataset \(\mathcal{D}_c\), training procedure, and model updates \(\mathbf{w}_c^t\) submitted to the server at each round \(t\). 

In a targeted backdoor attack, the adversary focuses on a specific class, known as the \textit{victim} class, and poisons only samples from this class. Specifically, the adversary injects \textit{poisoned} samples into the client’s local dataset by applying a trigger \(\mathbf{t}\) to inputs belonging to the victim class and flipping their labels to a designated \textit{target} class \(y_{\text{target}}\). Formally, for a subset \(\mathcal{D}_{\text{poisoned}} \subset \mathcal{D}_{\text{victim}}\), each input is modified as:
\begin{equation}
    \mathcal{D}_{\text{poisoned}} = \{(\mathbf{x}_i + \mathbf{t}, y_{\text{target}}) \mid (\mathbf{x}_i, y_i) \in \mathcal{D}_{\text{victim}}\},
\end{equation}
where \(\mathcal{D}_{\text{victim}}\) denotes the set of samples from the victim class.
The proportion of poisoned samples relative to the total number of clean samples is referred to as the \textit{Data Poisoning Rate} (DPR). Additionally, the adversary may alter the client's update \(\mathbf{w}_c^t\) before submission to the server, embedding adversarial changes that facilitate the backdoor's propagation to the global model. For convenience, definitions of all acronyms used throughout the paper can be found in \Cref{tab:glossary}.

\begin{table}[ht]
    \footnotesize
    \centering
    \begin{tabular}{ll|ll}
    \toprule
    \textbf{Acronym} & \textbf{Definition} & \textbf{Acronym} & \textbf{Definition} \\
    \midrule
    DPR & Data Poisoning Rate & MTA & Main Task Accuracy \\
    MCR & Malicious Client Ratio & ASR & Attack Success Rate \\
    \bottomrule
    \end{tabular}
    \caption{Glossary of acronyms used throughout the paper.}
    \label{tab:glossary}
\end{table}

\section{Limitation of Prior FL Backdoor Defenses}
\label{sec:existing_finicky}

Several works have proposed defenses against poisoning attacks in FL \cite{zhang2023flip, nguyen2022flame, blanchard2017machine, yin2018byzantine,fung2018mitigating,cao2021fltrust,wang2022flare,pillutla2022robust,shejwalkar2021manipulating}. These works usually focus on a specific learning configuration, for instance setting a specific learning rate and batch size for clients. While demonstrating superior performance in a particular setup is useful, it provides no guarantees about how well the defense would work in other valid learning configurations, \revision{and thus its sensitivity to a change in configurations}. We begin with an exploration of some of these learning configurations (\Cref{sec:fl_setup_matters}) and observe that there exist several equally-valid learning configurations where the model achieves acceptable MTA and the adversary's objective is preserved, thus making them all equally valid FL configurations. However, we find that all existing defenses we evaluated are effective only for a subset of these valid configurations and no defense is resilient across all configurations (\Cref{sec:defenses_fail_across_configs}).

\subsection{Learning Configuration Matters for Attack Success}
\label{sec:fl_setup_matters}

\begin{figure}[h!]
    \includegraphics[width=.98\linewidth]{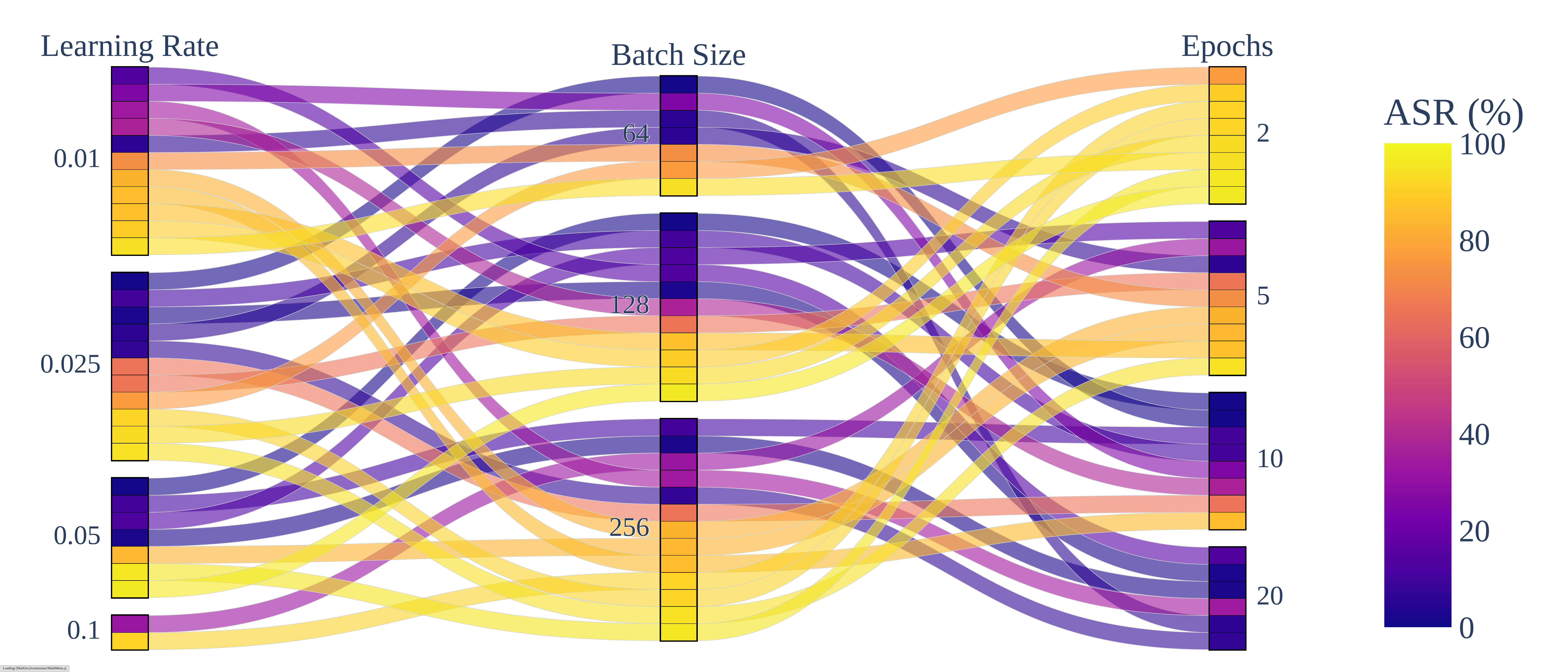}
    \caption{Visualizing the impact of the learning configuration (learning rate, batch size, and number of local epochs) on ASR, for 1.25\% DPR and 20\% MCR, for CIFAR-10 with IID data. We only visualize configurations with MTA $\geq80\%$. The attack is successful on multiple configurations (in yellow).}
    \label{fig:fl_setup_impact}
\end{figure}

To understand the impact of the learning configuration on model robustness and adversarial susceptibility, we conduct a grid-search analysis over key hyperparameters, varying the client's learning rate, batch size, and number of epochs on the CIFAR-10 dataset.
Our findings in \Cref{fig:fl_setup_impact} reveal that small changes in the learning configuration can significantly alter the model’s vulnerability to backdoor attacks.

\subsubsection{Impact of learning rate}

Lower learning rates (\eg 0.01, 0.025) allow for smaller, more gradual updates to the model parameters, making it easier for adversarial objectives to be embedded into the global model with minimal deviation from benign updates, resulting in a high ASR. In contrast, higher learning rates (\eg 0.1) cause larger, less stable updates that disrupt the optimization process, leading to degraded performance, with large drops in MTA and ASR alike.

\subsubsection{Impact of batch size}

Larger batch sizes (\eg 256) reduce variability in local updates, enabling high MTA and ASR across learning rates, making them more vulnerable to backdoor attacks. Smaller batch sizes (\eg 64) increase update variability, requiring very low learning rates (\eg 0.01) for effective backdoor injection.

\subsubsection{Impact of training epochs}

Fewer local epochs (\eg 2 or 5) lead to shallow optimization, reducing distinctions between benign and adversarial updates, making backdoor attacks more effective. Training with more local epochs (\eg 10 or 20) induces a \textit{polarization} effect, where clients' local models become more tightly aligned with their respective datasets. This stronger alignment reduces the influence of malicious updates during aggregation, as the backdoor signal becomes diluted and less effective at propagating into the global model. This polarization is even stronger when high LRs are used in conjunction with more training epochs, as model's MTA also suffers.

\Cref{tab:fl_setup_exps} highlights a range of configurations that we term the ``\textbf{danger zones}"---settings where the FL system achieves both high MTA (above 80\%) and adequately high ASR (above 85\%). These configurations are of great interest because they strike a balance between utility and vulnerability; accurate for legitimate tasks while remaining highly susceptible to targeted backdoor attacks. This highlights the critical need for a defense to be resilient across multiple learning configurations, ensuring robustness regardless of the setup and achieving true \textit{learning configuration independence}. This is particularly desirable because relying on a single configuration, even if effective, is not always feasible in practice. Various constraints, such as the batch size supported by a specific device, the number of epochs a client is willing to commit to, or other resource limitations, can dictate configurations in practice. Therefore, a defense mechanism that can adapt to different setups without compromising its efficacy is essential for practical and widespread deployment.

\subsection{Failure Modes of Prior Defenses}
\label{sec:defenses_fail_across_configs}

Based on our analysis, we identify 10 learning configurations where the attack is stealthy (minimal impact on MTA) and highly successful (has ASR higher than 85\%), as given in \Cref{tab:fl_setup_exps}. For our evaluation we consider key, prominent defense methods which offer diverse strategies to combat backdoor attacks in FL. These defense strategies include coordinate-based approaches like Median \cite{yin2018byzantine} and Multi-Krum \cite{blanchard2017machine}, trust-based methods such as FLTrust \cite{cao2021fltrust}, reputation-based schemes like FoolsGold \cite{fung2018mitigating} and FLARE \cite{wang2022flare}, anomaly detection frameworks like FLAME \cite{nguyen2022flame} and local adversarial training methods like FLIP \cite{zhang2023flip}. For more details on these defenses and other related works, see \Cref{sec:experiments} and \Cref{sec:related_work}.

\begin{table}[h]
    \centering
    \small
    \begin{tabular}{llcc|cc}
    \toprule
    \textbf{Config} & \textbf{LR} & \textbf{BS} & \textbf{Epochs} & \textbf{MTA (\%)} & \textbf{ASR (\%)} \\
    \midrule
    C1 & 0.05 & 128 & 2 & 85.08 & 96.5 \\
    C2 & 0.05 & 256 & 2 & 86.29 & 95.8 \\
    C3 & 0.025 & 256 & 5 & 88.33 & 94.8 \\
    C4 & 0.01 & 64 & 2 & 87.03 & 93.9  \\
    C5 & 0.025 & 128 & 2 & 86.65 & 92.9 \\
    C6 & 0.025 & 256 & 2 & 82.97 & 91.3 \\
    C7 & 0.1 & 256 & 2 & 85.55 & 91.1 \\
    C8 & 0.01 & 128 & 2 & 84.47 & 89.7 \\ 
    C9 & 0.01 & 128 & 5 & 89.04 & 86.6\\
    C10 & 0.01 & 256 & 10 & 87.97 & 85.6 \\
    \bottomrule
    \end{tabular}
    \caption{Client FL configurations for successful stealthy attacks on CIFAR-10 \ie cases with MTA $\geq 80\%$ and ASR $\geq 85\%$. }
    \label{tab:fl_setup_exps}
\end{table}
Evaluation setups for these defenses are often inconsistent, with significant variation in client learning configurations such as learning rate, batch size, and the number of local training epochs. For instance, some works (e.g., FLTrust) specify fixed learning rates and batch sizes, while others (e.g., Multi-Krum, FLAME) provide little to no details about the client training setup, focusing instead on aggregation logic. This lack of standardization in evaluation protocols raises concerns about the reproducibility and generalizability of reported results. A detailed comparison of these evaluation setups is provided in \Cref{app:baseline_details}.

\begin{figure}[h]
    \includegraphics[width=.9\linewidth]{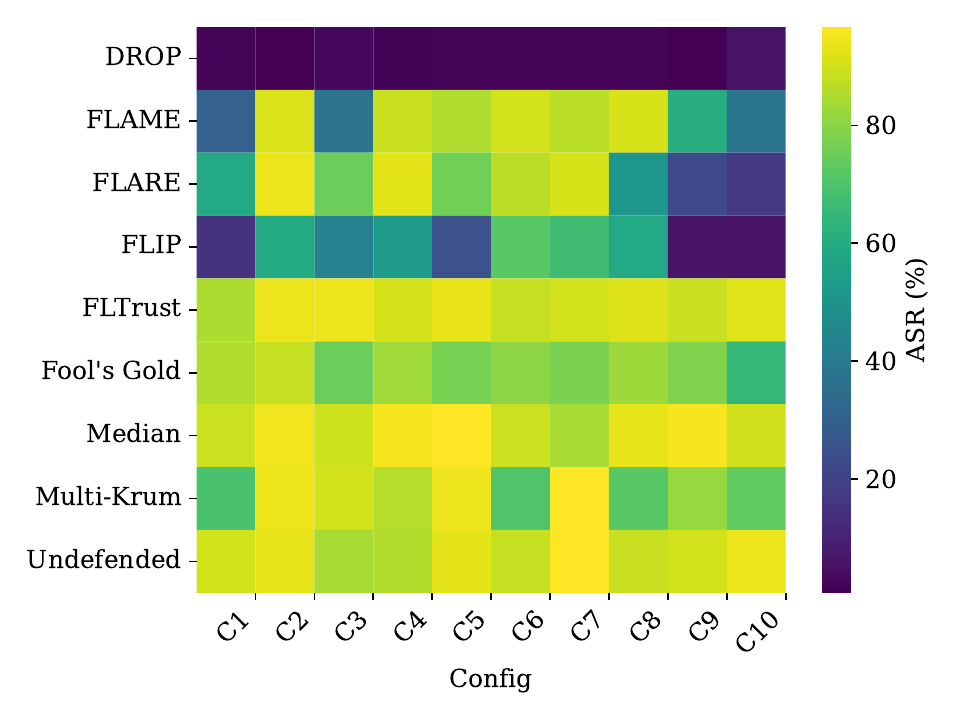}
    \caption{ASR (\%) for our defense (DROP) and various existing defenses for 10 FL configurations (1.25\% DPR, 20\% MCR) where stealthy attacks are possible. No existing defense provides consistent protection across all configurations.}
    \label{fig:baselines_heatmap}
\end{figure}

We find that these approaches often struggle against targeted backdoor attacks due to the attack's stealthy nature (\Cref{fig:baselines_heatmap}). Median and Multi-Krum fail to exclude the adversary’s updates since the poisoned gradients remain close to the statistical norm, while FLTrust struggles to identify the malicious contributions due to the minimal deviation from expected update patterns. Similarly, FoolsGold proves largely ineffective because the adversary’s contributions exhibit insufficient variability across rounds, making them difficult to detect. FLAME performs reasonably well in certain configurations but fails against attacks that exploit low learning rates. Among the defensive baselines, FLIP is the only one that slightly reduces the ASR, though its mitigation is insufficient to provide robust protection.
Existing defenses thus have a severe limitation when it comes to stealthy backdoors: \textbf{their resilience is sensitive to the specific FL learning configuration.} \revision{In contrast, our proposed defense (which we describe next) generalizes to all learning setups and provides consistent protection.}

\section{Our Defense: DROP}
\label{sec:proposed_defense}

\begin{figure*}[htbp]
    \centering
    \includegraphics[width=0.9\textwidth]{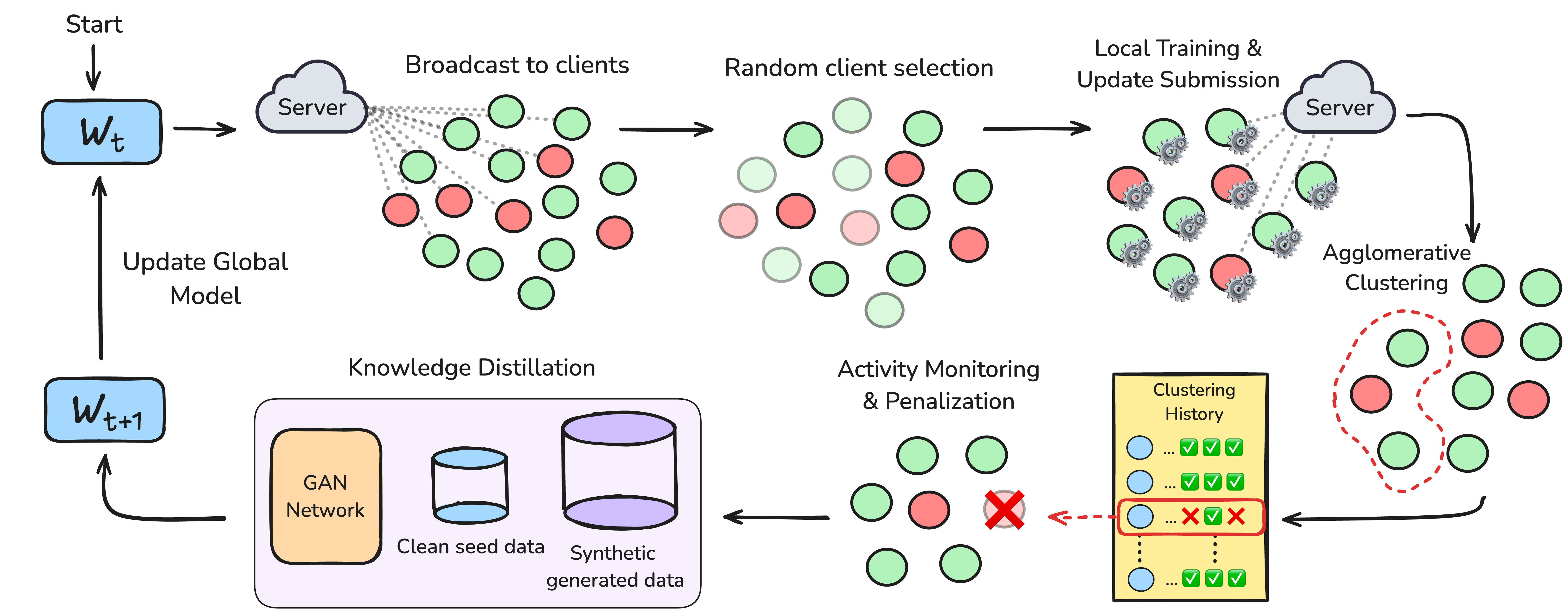}
    \caption{Overview of the proposed DROP defense. Each round \( t \) begins with the server broadcasting the global model to all clients and selecting a subset for local training, which may include both benign (green) and malicious (red) clients. After updates are submitted, DROP employs: (1) \textbf{Agglomerative Clustering} to detect anomalous updates, (2) \textbf{Activity Monitoring \& Penalization} to track and penalize suspicious clients, and (3) \textbf{Knowledge Distillation}, where a GAN-generated synthetic dataset and client logits guide the distillation of the global model. The final model \( \mathbf{w}_{t+1} \) serves as the global model for round \( t+1 \).}
    \label{fig:drops} 
\end{figure*}

From our analysis, it is clear that the learning configuration of a FL system plays a pivotal role in the success of targeted backdoor attacks. Certain configurations exhibit inherent vulnerabilities, allowing adversarial updates to bypass detection and achieve high ASR. Moreover, the performance of existing defenses shows significant variability across different configurations, revealing their lack of robustness across learning setups. These findings emphasize the urgent need for a \textit{universal} defense mechanism that is resilient against a wide array of attack strategies and remains agnostic to the underlying learning configuration. 

To address this challenge, we introduce \textbf{DROP} (\textit{\textbf{D}istillation-based \textbf{R}eduction \textbf{O}f \textbf{P}oisoning}), a federated framework designed to counter three critical adversarial scenarios:
\begin{itemize}
    \item aggressive adversaries that may resort to high-DPR attacks,
    \item diverse MCRs  within the federation, and
    \item stealthy, low-DPR attacks that exploit specific learning configurations.
\end{itemize}
These scenarios each present unique challenges that require a cascade of countermeasures, as summarized in the following sections. An overview of our approach is given in \Cref{fig:drops}.

The first countermeasure, \textit{Agglomerative Clustering} (\Cref{subsec:agglo_clustering}), identifies and isolates malicious updates by clustering submitted models based on their pairwise Euclidean distances.
This process is particularly effective against high-DPR attacks, where malicious updates deviate significantly from benign updates.  
The second countermeasure, \textit{Activity Monitoring} (\Cref{subsec:activity_monitoring}), tracks client behavior across training rounds. By maintaining a reputation score based on prior clustering results, this mechanism penalizes clients frequently flagged as suspicious and reduces their influence in future aggregations. This reputation-based approach ensures robustness against per round benign-to-malicious client ratio fluctuations, where the proportion of malicious clients can vary widely across rounds due to random client selection.  

The third and final countermeasure, \textit{Knowledge Distillation} (\Cref{subsec:kd}), addresses the most challenging class of attacks: stealthy, low-DPR strategies. These attacks encode adversarial objectives subtly into model updates, making them indistinguishable from benign updates and enabling them to bypass clustering schemes. To neutralize these residual adversarial signals, we use synthetic data generated by a GAN trained via logit-driven distillation to produce a \textit{cleansed} version of the global model. This process ensures robustness against low-DPR attacks and renders the defense agnostic to learning configurations. 

\revision{DROP is the first FL defense that leverages knowledge distillation to remove stealthy poisoning attacks. DROP's main innovation is in its architecture that leverages these three components to counteract a wide range of adversarial strategies.} \revision{In particular, through this novel set of countermeasures, DROP 
provides resilience against aggressive, high data poisoning rates, adaptability to diverse malicious client ratios, and robustness against stealthy, low data poisoning rates, across learning configurations.}  

Next, we provide a detailed explanation of the design and motivation behind each component. The entire DROP algorithm is presented in Algorithm \ref{alg:drop}. 

\begin{algorithm}
\caption{DROP}
\label{alg:drop}
\SetKwInput{kwGlobals}{Globals input}
\SetKwFunction{FMain}{DROP}
\SetKwFunction{FClustering}{AgglomerativeClustering}
\SetKwFunction{FActivity}{UpdateActivity}
\SetKwFunction{FScore}{score}
\SetKwFunction{FDistillation}{KD}
\SetKwProg{Fn}{Function}{:}{}
\SetKwFor{ForEach}{for each}{do}{end}

\kwGlobals{initial model parameters $w_0$, total training rounds $T$, total client set $\mathcal{N}$, ban threshold $\tau_b$, generator $G$, clean seed set $\mathcal{D}_{\text{clean}}$, knowledge-distillation $KD$}

\For{each round $t \in \{1, \dots, T\}$}{
    \tcp{\textcolor{blue}{Step 1: Training and Update Collection}}
    \ForEach{client $c \in \mathcal{C}_t \subset \mathcal{N}$}{
        $w_c^t \leftarrow \text{ClientLocalTraining}(w_{t-1})$ \tcp*[r]{\textcolor{blue}{Client $c$ trains locally and returns update}}
    }
    
    \tcp{\textcolor{blue}{Step 2: Agglomerative Clustering}}
    $\mathcal{C}_{\text{b}}, \mathcal{C}_{\text{s}} \leftarrow \FClustering(\{w_c^t : c \in \mathcal{C}\})$ \tcp*[r]{\textcolor{blue}{Cluster client updates into \textbf{b}enign and \textbf{s}uspect groups}}
    
    \tcp{\textcolor{blue}{Step 3: Activity Monitoring}}
    \ForEach{client $c \in \mathcal{C}$}{
        \FActivity($c$) %
        \If{$c \in \mathcal{C}_{\text{b}} \land \FScore(c) \geq \tau_b$}{
            \tcp{\textcolor{blue}{Exclude $c$ if it exceeds penalty threshold}}
            $\mathcal{C}_{\text{b}} \leftarrow \mathcal{C}_{\text{b}} \setminus \{c\}$ %
        }
    }
    
    \tcp{\textcolor{blue}{Aggregate model updates from benign clients}}
    $w_t \leftarrow \text{Aggregate}(\{w_c^t : c \in \mathcal{C}_{\text{b}}\})$ %

    \tcp{\textcolor{blue}{Step 4: Logit-Driven Model Distillation}}
    $w_t \leftarrow \FDistillation(w_t, \{w_c^t : c \in \mathcal{C}_{\text{b}}\}, G, \mathcal{D}_{\text{clean}})$ %

    \tcp{\textcolor{blue}{Step 5: Broadcast the Cleansed Global Model}}
        \ForEach{client $c \in \mathcal{C}$}{
            \text{Send}($c, w_t$);
        }
}

\end{algorithm}

\subsection{Agglomerative Clustering}  
\label{subsec:agglo_clustering}  

The first line of defense in our system is \textit{Agglomerative Clustering} (AC) \cite{müllner2011modernhierarchicalagglomerativeclustering}, selected for its flexibility and hierarchical structure, which makes it well-suited for detecting subtle deviations caused by targeted backdoor attacks. Clustering is a widely adopted countermeasure in filtering-based FL frameworks \cite{nguyen2022flame, fung2018mitigating, baybfed, mesas, auror} due to its ability to identify dissimilarities within model updates, often resulting from adversarial strategies that embed backdoors aggressively with a high DPR. We chose AC over methods like K-means and HDBSCAN due to its lower sensitivity to hyper-parameter selection and its dynamic merging process, which eliminates the need to pre-specify the number of clusters. Unlike K-means, which can be sensitive to initial centroid selection and outliers, AC provides more consistent results by iteratively merging clusters based on similarity. Similarly, while HDBSCAN \cite{campello2013density} relies on density-based assumptions that require careful tuning of hyper-parameters, AC operates without such assumptions, making it more robust in diverse FL scenarios.

The use of \textit{Ward linkage} \cite{Ward01031963} further enhances AC's effectiveness by ensuring that intra-cluster variance is minimized during the merging process. Among the various linkage methods available for AC, Ward linkage is particularly effective for FL tasks, as it prioritizes clusters with low internal variance, which is crucial for distinguishing fine-grained and stealthy backdoors embedded within benign-looking updates. By capturing and separating subtle differences in high-dimensional weight spaces, Ward linkage ensures that adversarial updates are reliably isolated from legitimate ones.  

Let \(\mathcal{W} = \{\mathbf{w}_1, \mathbf{w}_2, \dots, \mathbf{w}_n\}\) denote the set of model updates from \(n\) participating clients in a given training round \(t\). Let $d(\mathbf{w}_i, \mathbf{w}_j) = \| \mathbf{w}_i - \mathbf{w}_j \|_2$ be the Euclidean distance between two model updates \(\mathbf{w}_i\) and \(\mathbf{w}_j\).
The clustering process starts with each client update \(\mathbf{w}_i\) being treated as a singleton cluster. At each step, the two clusters \(\mathcal{A}\) and \(\mathcal{B}\) with the smallest inter-cluster distance are merged. Using \textit{Ward linkage}, the distance between two clusters \(\mathcal{A}\) and \(\mathcal{B}\) is defined as the increase in total intra-cluster variance caused by merging the two clusters:
\begin{equation}
    d_{\text{Ward}}(\mathcal{A}, \mathcal{B}) = \frac{|\mathcal{A}| \, |\mathcal{B}|}{|\mathcal{A}| + |\mathcal{B}|} \, \| \boldsymbol{\mu}_{\mathcal{A}} - \boldsymbol{\mu}_{\mathcal{B}} \|_2^2,
\end{equation}
where \(|\mathcal{A}|\) and \(|\mathcal{B}|\) are the sizes (number of points) of clusters \(\mathcal{A}\) and \(\mathcal{B}\), and \(\boldsymbol{\mu}_{\mathcal{A}}\) and \(\boldsymbol{\mu}_{\mathcal{B}}\) are their respective centroids.

The hierarchical clustering process proceeds iteratively until a stopping criterion is met. In our case, the process halts upon forming exactly two predefined clusters, \(\{\mathcal{C}_b, \mathcal{C}_s\}\), which effectively capture natural groupings of benign and malicious model updates.
In the context of FL, it is assumed that benign client updates will cluster together due to the shared training objective, while adversarial updates will form separate, smaller clusters due to their deviation from normal update patterns.

Minimizing intra-cluster variance ensures that the clustering process captures subtle distinctions in high-dimensional weight spaces, where even minor variations can signal meaningful differences, such as those between adversarially perturbed weights and legitimate ones. The hierarchical structure produced by Ward linkage allows flexibility in examining update patterns at various levels of similarity, enabling dynamic control over how clusters are merged. This approach is particularly valuable in FL, where client updates are naturally noisy due to heterogeneous data distributions but can also be adversarially manipulated. By clustering updates, this method enables the system to separate and potentially eliminate outliers or manipulated model updates.
\goalbox{\underline{\textbf{Goal 1}}: The clustering component aims to eliminate malicious updates that deviate significantly from their benign counterparts in each training round, offering resilience against high data poisoning rate attacks.}\label{goal1}

\subsection{Activity Monitoring}
\label{subsec:activity_monitoring}

The second line of defense, complementing the clustering component, is an \textit{activity monitoring mechanism}, which tracks the behavior of participating clients over the course of training. 

This mechanism addresses a limitation of clustering-based defenses in federated learning, which assume that the number of malicious updates in a round is smaller than the number of benign ones. In real-world scenarios, however, client selection is random, and the defender has no prior knowledge of the number of malicious clients in any given round. This randomness can lead to rounds where adversarial clients outnumber benign ones, causing misidentification of the smaller cluster.
Prior works often assume an upper bound on the proportion of malicious clients (typically less than 50\%), but enforcing this requires low MCRs or artificially controlled client sampling  to maintain a fixed benign-to-malicious client ratio across rounds \cite{zhang2023flip}, limiting practical applicability.
It is important to note that although the total proportion of adversarial clients in the federation (MCR) remains fixed, the actual ratio of malicious clients in any given round can fluctuate due to the random selection of clients.

The maximum tolerable MCR of a defense—the fraction of clients in the federation that can be adversarial while still allowing the defense to mitigate an attack—depends on the total number of clients in the federation, \(N\), and the number of clients sampled in each round, \(C\). 
Under a random sampling strategy, \(C\) clients are drawn uniformly from the total \(N\) clients. Let $\rho$ be the MCR \ie proportion of malicious clients in the entire federation, so the expected number of malicious clients ($\mathcal{M})$ in a round is \(\mathbb{E}[\mathcal{M}] = \rho \cdot C\). By modeling the selection of clients as a Binomial distribution, we can compute a lower bound on the probability of malicious clients outnumbering benign ones in any given round as:
\begin{align}
    P\left(\mathcal{M} \geq \frac{C}{2}\right) \geq  1 - \left(4\rho(1-\rho)\right)^{\frac{C}{2}}.
\end{align}
For a derivation of the above, please see \Cref{app:bound_analysis}.
This bound implies that as \(C\) increases, the likelihood of selecting a disproportionately large number of malicious clients diminishes exponentially. Conversely, when \(C\) is small,
the probability that malicious clients outnumber benign clients within the selected subset increases. 
For example, consider an FL system with \(N = 100\) total clients and a MCR of 40\% (\(\rho = 0.4\)). If \(C = 20\) clients are randomly selected in a given round, the expected number of malicious clients in the subset is \(\mathbb{E}[\mathcal{M}] = \rho \cdot C = 0.4 \cdot 20 = 8\).
However, with the inequality above we can see the probability of selecting more than 10 malicious clients can be \underline{non-trivial} ($\approx 0.34$), potentially resulting in a subset where malicious clients constitute more than 50\% of the selected participants (\(M > C/2\)). In fact, the bound suggests that in about a third of the FL training rounds, the malicious clients will be the majority. 

To address scenarios where the number of malicious clients exceeds the benign ones in certain rounds, we propose a reputation-based mechanism. This approach tracks client behavior across rounds, penalizing suspicious clients identified by the clustering component (\Cref{subsec:agglo_clustering}) and gradually reducing their influence over the global model. By doing so, our method dynamically adjusts to varying levels of malicious participation without relying on rigid assumptions about MCR or artificially controlling client sampling. For instance, FLIP~\cite{zhang2023flip} enforces a fixed per-round MCR by '\textit{randomly}' selecting 10 clients per round, ensuring exactly 4 adversaries and 6 benign clients, thus artificially tampering with the randomness of client selection.
By incorporating a reputation system, our defense mitigates errors caused by transient imbalances where adversarial clients may temporarily outnumber benign ones. Over multiple rounds, as long as the overall MCR remains below 50\%, the mechanism ensures that benign clients are correctly identified and retained more frequently than adversarial clients. This design helps prevent long-term accumulation of adversarial influence, even in rounds where clustering misclassifications may occur.

Our reputation-based mechanism employs a penalty and reward system that essentially tracks the trustworthiness of each client, enabling the server to mitigate persistent malicious activity while accounting for potential false positives. Specifically, by maintaining a cumulative penalty score \(\pi(c)\) for each client \(c\), the server can distinguish between clients with occasional false-positive detections and those consistently submitting suspicious updates. This ensures that benign clients incorrectly flagged as suspicious in a few rounds do not face permanent exclusion from the system.

\shortsection{Calculating Penalty Scores}
We begin by initializing the penalty score \(\pi(c) = 0\) for every client. As training proceeds, this score is dynamically updated based on the clustering results from each round.
During each training round \( t \), the clustering component (\Cref{subsec:agglo_clustering}) determines whether a client's update is benign or suspicious by assigning the client with a penalty score:
\begin{align}
    \pi(c) =
    \begin{cases}
    \pi(c) + p, & \text{if } c \in \mathcal{C}_s\\
    \max(0, \pi(c) - r),              & \text{otherwise}
    \end{cases}
\end{align}
\newline
Essentially, if a client's update is flagged as \textit{suspicious} (\( c \in \mathcal{C}_s \)), its penalty score is increased by a constant penalty value \( p \). Otherwise, it is reduced by a constant reward value \( r \), ensuring the score remains non-negative. Capping the penalty score at zero prevents potentially malicious clients from gaining undue advantage in cases where they are repeatedly flagged as benign due to false positives. $p$ and $r$ are hyper-parameters determined by the server.

The server uses these scores to regulate client participation. Clients with a clean history \ie consistently flagged as benign (\(\pi(c) = 0\)) are deemed trustworthy and allowed to contribute to the global model, whereas clients with a record of suspicious update submissions (\(\pi(c) > 0\)) are restricted from participating in the aggregation process. However, depending on the server administration policy, the penalty system could be even stricter. Clients that accumulate excessive penalty points due to repeated suspicious updates are permanently \textit{blacklisted} from contributing in subsequent rounds:

Optionally, the server could permanently ban any client whose penalty score exceeds a predefined threshold.
\goalbox{\underline{\textbf{Goal 2}}: The activity monitoring component helps prevent malicious actors (based on their accumulated penalty points) from poisoning the global model during rounds in which the malicious outnumber the benign participants.}\label{goal2}

\subsection{Knowledge Distillation}
\label{subsec:kd}

Having tackled high-DPR and dynamic-MCR attacks, the most challenging category to defend against is low-DPR attacks.
These attacks are difficult to detect due to their stealthy nature, where adversarial objectives are subtly embedded into model weights, making the updates nearly indistinguishable from benign ones. This smooth encoding allows them to bypass traditional clustering-based defenses that rely on statistical or geometric properties of the updates. To counter this, DROP introduces a knowledge distillation-based cleansing mechanism that neutralizes residual adversarial signals in the global model using clean, synthetic data, effectively mitigating the impact of stealthy backdoor updates while preserving the model’s utility for legitimate tasks.

This approach leverages model stealing attacks, such as the \textit{MAZE} model stealing framework \cite{kariyappa2021maze}, which achieves state-of-the-art performance in extracting machine learning models under black-box conditions. The goal here is not to engage in model stealing as traditionally intended, but to clone the model without introducing any malicious behavior from compromised clients. \textit{MAZE} combines two key components: query synthesis and knowledge distillation. Knowledge Distillation \cite{hinton2015distillingknowledgeneuralnetwork} is a technique where a "student" model learns to approximate the output logits of a larger "teacher" model, enabling the transfer of knowledge while preserving essential decision boundaries. A generative adversarial network (GAN) \cite{radford2016unsupervisedrepresentationlearningdeep} synthesizes inputs to probe a target model's decision boundaries, and knowledge distillation transfers the target model's behavior to a surrogate model trained on these synthetic queries. By minimizing reliance on labeled data, \textit{MAZE} efficiently approximates the target model's functionality.

In our approach, we adapt and extend the \textit{MAZE} framework to take advantage of the white-box access available to the server in FL, as the server owns the global model. This adjustment eliminates the need for zeroth-order gradient estimation used in black-box settings and allows for direct backpropagation, improving both efficiency and accuracy. To ensure high-quality synthetic data, we initialize the GAN with a set of \( n \) clean samples, denoted as \( \mathcal{D}_\text{clean} = \{\mathbf{x}_i \}_{i=1}^n \) in step 4 of Algorithm \ref{alg:drop}, which helps align the generated queries with the original data distribution. \revision{These samples are drawn from the same data distribution as that of the federation clients and serve as a lightweight initialization prior for the generator.} Our approach requires only a small set of clean samples—typically less than or equal to the size of a single client’s dataset—ensuring that the additional data requirement remains realistic and practical for deployment in real-world FL scenarios.

In a training round $t$, instead of relying solely on the global model’s logits, we aggregate logits from client updates that pass the filtering process (\ie $\mathcal{C}_b$). Let the logit output from the global model be \( \mathbf{z}_t = f(\mathbf{x}; \mathbf{w}_t) \), where \( \mathbf{w}_t \) represents the global model parameters at round \( t \). Similarly, let the logit outputs from benign client models \( \mathcal{C}_b \subseteq \mathcal{C}_t \) be:
\begin{equation}
    \mathbf{z}_{t}^c = f(\mathbf{x}; \mathbf{w}_t^c), \quad \forall c \in \mathcal{C}_b.
\end{equation}
We compute the ensemble logits \( \mathbf{\bar{z}}_t \) as the average of the logits from benign clients:
\begin{equation}
    \mathbf{\bar{z}}_t = \frac{1}{|\mathcal{C}_b|} \sum_{c \in \mathcal{C}_b} \mathbf{z}_{t}^c.
\end{equation}

The generator \( G \) uses the ensemble logits \( \mathbf{\bar{z}}_t \) as feedback to synthesize new queries \( \mathcal{D}_{\text{synthetic}} \). These synthetic queries, denoted as \( \mathbf{x}_{\text{gen}} \), are used to guide a \textit{clone network}, which serves as a cleansed version of the global model. The clone network, parameterized by \( \mathbf{w}^{\text{clone}}_t \), is trained to align its predictions with the ensemble logits. Instead of the KL divergence used in the original \textit{MAZE} framework, we employ the \(\ell_1\)-loss, which provides more stable performance \cite{truong2021data}. The training objective for the clone network is defined as:
\begin{equation}
    \mathcal{L}_{\text{distill}} = \mathbb{E}_{\mathbf{x} \sim \mathcal{D}_{\text{synthetic}}} \left[ \| \mathbf{\bar{z}}_t - \mathbf{z}_t^{\text{clone}} \|_1 \right],
\end{equation}
where \( \mathbf{z}_t^{\text{clone}} = f(\mathbf{x}; \mathbf{w}^{\text{clone}}_t) \) represents the logit predictions of the clone network on synthetic data \( \mathbf{x} \).

The key difference between this approach and the original MAZE framework is that, instead of distilling knowledge from a single victim model, we distill knowledge from the aggregated logits of benign clients. The intuition behind this approach is that the ensemble logits \( \mathbf{\bar{z}}_t \), derived from benign client models, act as a \textit{consensus signal} to overwrite any adversarial influence introduced by poisoned updates. By aligning the clone network’s behavior with the aggregated benign logits, the distillation process effectively cleanses the model of any stealthy backdoor or poisoned behavior that might have evaded detection in previous defense layers. 
\goalbox{\underline{\textbf{Goal 3}}: The collective logit-driven knowledge distillation framework ensures that any subtle adversarial updates which have made their way to the global model are neutralized, restoring the global model’s robustness and reliability.}\label{goal3}

\subsection{DROPlet: a Lightweight, Scalable Defense Mechanism}  
\label{subsec:droplet}  

To provide a faster, more resource-efficient alternative to DROP, we introduce \textit{DROPlet}, a lightweight, server-side plugin designed for easy deployment and high scalability.
Unlike DROP, DROPlet omits the knowledge distillation component and focuses solely on agglomerative clustering and activity monitoring to mitigate adversarial influences with minimal overhead.  

DROPlet is task-agnostic, with no assumptions on the underlying data modality, making it applicable to various FL tasks. This versatility allows DROPlet to function as a general-purpose defense mechanism in diverse FL deployments. By eliminating the computationally intensive knowledge distillation step, it achieves faster round completion times, making it particularly suitable for large-scale FL systems where computational efficiency is crucial.  

Despite its lightweight design, DROPlet offers robust protection by leveraging clustering to detect anomalies and activity monitoring to penalize malicious clients. However, it may not be effective against stealthy, low-DPR attacks.
Nevertheless, DROPlet remains a competitive baseline for FL defense evaluations, offering significant protection while being faster and easier to deploy compared to classical methods like Multi-KRUM and Median. 

\section{Experiments}
\label{sec:experiments}

We conduct extensive experiments under a wide range of adversarial conditions and data distributions. Our evaluations focus on various learning configurations of interest, as discussed in \cref{sec:defenses_fail_across_configs}. We also vary \revision{attack strategies}, DPRs, MCRs, and use both independent and identically distributed (IID) and non-IID data distributions, ensuring a comprehensive assessment of DROP in realistic FL settings.

\shortsection{Datasets and Model Architecture} 
We evaluated our defense using three standard FL datasets: CIFAR-10, \revision{CINIC-10}, and EMNIST.
CIFAR-10 \cite{krizhevsky2009learning} contains 60,000 32x32 color images across 10 diverse classes.
\revision{CINIC-10 \cite{darlow2018cinic} is an extension of CIFAR-10, incorporating additional images from ImageNet \cite{imagenet} to increase dataset size and variability. It consists of 270,000 images distributed across the same 10 classes as CIFAR-10, providing a more challenging classification task with greater intra-class diversity.}
EMNIST \cite{cohen2017emnistextensionmnisthandwritten} extends MNIST \cite{mnist} with 62 handwritten character and digit classes, making it suitable for FL classification tasks. We use a ResNet-18 architecture \cite{he2015deepresiduallearningimage} for all experiments.

\shortsection{Data Distribution} 
We evaluate our defense on IID and across various levels of non-IID data to reflect the heterogeneity typically observed in real-world FL deployments. To control the degree of non-IID-ness, we partition the global dataset using a Dirichlet distribution $Dir(\alpha)$ \cite{li2021federatedlearningnoniiddata} (see \Cref{sec:non_iid}) and use the distributional parameter $\alpha$ to vary the degree of non-IID-ness.

\shortsection{Federation Settings} The federation consists of 50, 100, and 150 clients for CIFAR-10, EMNIST, and \revision{CINIC-10} respectively. 50\% clients are randomly selected to participate each training round for CIFAR-10 and EMNIST \revision{while for CINIC-10, 20\% clients are randomly selected per training round.}
Unlike prior works \cite{blanchard2017machine,zhang2023flip} that assume a fixed number of adversarial clients in every round $t$, our model allows for the number of adversarial clients in \(\mathcal{C}_t\) to vary.

\begin{table*}[h]
    \centering
    \resizebox{\textwidth}{!}{
        \begin{tabular}{l|c|ccc|ccc||c|cc|cc||c|cc|cc}
        \toprule
        \multirow{2}{*}{\textbf{Defense}} & \multicolumn{7}{c}{CIFAR-10} & \multicolumn{5}{c}{CINIC-10} & \multicolumn{5}{c}{EMNIST} \\
         & \multicolumn{4}{c}{MTA (\%)} & \multicolumn{3}{c}{ASR (\%)} & \multicolumn{3}{c}{MTA (\%)} & \multicolumn{2}{c}{ASR (\%)} & \multicolumn{3}{c}{MTA (\%)} & \multicolumn{2}{c}{ASR (\%)} \\
         \cline{2-18}
         \multicolumn{1}{r|}{DPR (\%)} &  NA & 1.25 & 2.5 & 5 & 1.25 & 2.5 & 5 & NA & 1.25 & 2.5 & 1.25 & 2.5 & NA & 1.25 & 2.5 & 1.25 & 2.5 \\
        \midrule
        Undefended & 87.56 & 87.55 & 87.32 & 87.40 & 85.5 & 96.6 & 98.4 & 74.71 & 74.61 & 74.49 & 94.6 & 98.65 & 89.01 & 89.20 & 89.11 & 98.50 & 97.75 \\
        \hline
        FLAME & 85.12 & 84.86 & 85.00 & 85.03 & 88.7 & 96.8 & 99.4 & 72.93 & 73.16 & 72.77 & 96.63 & 98.6 & 88.02 & 87.83 & 88.0 & 5.0 & 12.75 \\
        FLARE & 85.29 & 85.06 & 81.08 & 84.41 & 92.5 & 75.8 & 3.5 & 71.07 & 72.18 & 71.09 & 96.71 & 93.02 & 88.82 & 89.07 & 88.92 & 96.00 & 84.50 \\
        FLIP & 81.25 & 83.09 & 81.53 & 82.88 & 53.4 & 73.4 & 48.80 & 67.73 & 70.08 & 69.60 & 80.98 & 38.82 & 88.31 & 88.53 & 88.32 & 46.50 & 48.75 \\
        FLTrust & 87.68 & 87.93 & 87.52 & 87.62 & 90.4 & 96.7 & 98.1 & 74.46 & 74.74 & 74.54 & 90.83 & 98.72 & 89.22 & 89.19 & 89.18 & 97.0 & 90.0 \\
        Fool's Gold & 83.59 & 83.85 & 83.76 & 83.92 & 83.0 & 92.6 & 94.3 & 70.25 & 70.05 & 69.96 & 89.33 & 95.29 & 87.92 & 87.85 & 87.80 & 77.0 & 46.50 \\
        Median & 87.07 & 87.48 & 87.33 & 87.21 & 95.3 & 96.3 & 98.8 & 73.29 & 73.49 & 73.38 & 97.6 & 98.87 & 88.94 & 89.20 & 89.20 & 96.75 & 96.25 \\
        Multi-KRUM & 85.66 & 85.96 & 86.00 & 85.87 & 86.0 & 97.5 & 98.6 & 73.61 & 73.57 & 73.58 & 76.35 & 99.74 & 88.62 & 88.70 & 88.58 & 78.75 & 5.0 \\
        \hline
        DROPlet (ours) & 87.33 & 87.49 & 87.05 & 87.20 & 91.3 & \textbf{0.1} & 0.9 & 74.62 & 74.05 & 74.27 & 15.96 & \textbf{0.48} & 89.15 & 88.98 & 88.96 & 1.50 & 1.00 \\
        DROP (ours) & 76.07 & 76.05 & 76.53 & 75.18 & \textbf{1.3} & 1.1 & \textbf{0.5} & 65.4 & 65.65 & 65.72 & \textbf{1.22} & 1.44 & 89.32 & 88.62 & 89.13 & \textbf{0.75} & \textbf{1.00} \\
        \bottomrule
        \end{tabular}
    }
    \caption{MTA (\%) and ASR (\%) metrics under no attack (NA) and at different Data Poisoning Rates (DPR) for the CIFAR-10, \revision{CINIC-10} and EMNIST datasets. The MCR is fixed at 20\%. DROP achieves low ASR for all DPR, while existing defenses fail to mitigate the attack. The lightweight variant DROPlet achieves higher MTA than DROP and is robust in all cases, except for some extremely stealthy attacks (1.25\% DPR, CIFAR-10). The learning configurations are C4 for CIFAR-10 (same for CINIC-10) and C5 for EMNIST. Refer to \cref{tab:fl_setup_exps} and \cref{tab:fl_setup_exps_emnist} for learning configuration details.}
    \label{tab:fl_vary_dpr}
    
\end{table*}

\begin{table*}[h]
    \centering
    \resizebox{\textwidth}{!}{
        \begin{tabular}{l|l|ccc|ccc||cc|cc||cc|cc}
        \toprule
        \multirow{2}{*}{\textbf{Attack}} & \multirow{2}{*}{\textbf{Defense}} & \multicolumn{5}{c}{CIFAR-10} & \multicolumn{4}{c}{CINIC-10} & \multicolumn{4}{c}{EMNIST} \\
        & & \multicolumn{3}{c}{MTA (\%)} & \multicolumn{3}{c}{ASR (\%)} & \multicolumn{2}{c}{MTA (\%)} & \multicolumn{2}{c}{ASR (\%)} & \multicolumn{2}{c}{MTA (\%)} & \multicolumn{2}{c}{ASR (\%)} \\
        \cline{3-16}
        & \multicolumn{1}{r|}{DPR (\%)} & 1.25 & 2.5 & 5 & 1.25 & 2.5 & 5 & 1.25 & 2.5 & 1.25 & 2.5 & 1.25 & 2.5 & 1.25 & 2.5 \\
        \midrule
        \multirow{4}{*}{BadNets} & Undefended & 87.55 & 87.32 & 87.40 & 85.5 & 96.6 & 98.4 & 74.61 & 74.49 & 94.6 & 98.65 & 89.20 & 89.11 & 98.50 & 97.75 \\
        & FLIP & 83.09 & 81.53 & 82.88 & 53.4 & 73.4 & 48.80 & 70.08 & 69.60 & 38.82 & 88.53 & 88.53 & 88.32 & 46.50 & 48.75 \\
        & DROPlet (ours) & 87.49 & 87.05 & 87.20 & 91.3 & \textbf{0.1} & 0.9 & 74.05 & 74.27 & 15.96 & \textbf{0.48} & 88.98 & 88.96 & 1.50 & 1.00 \\
        & DROP (ours) & 76.05 & 76.53 & 75.18 & \textbf{1.3} & 1.1 & \textbf{0.5} & 65.65 & 65.72 & \textbf{1.22} & 1.44 & 88.62 & 89.13 & \textbf{0.75} & \textbf{1.00} \\
        \hline
        \multirow{4}{*}{Neurotoxin} & Undefended & 87.64 & 87.27 & 87.09 & 92.20 & 93.20 & 98.2 & 74.24 & 74.29 & 97.50 & 98.54 & 89.16 & 89.17 & 98.0 & 97.75 \\
        & FLIP & 82.76 & 83.18 & 82.04 & 53.0 & 84.40 & 73.4 & 69.83 & 69.87 & 24.52 & 77.27 & 88.52 & 88.49 & 41.25 & 10.75 \\
        & DROPlet (ours) & 87.10 & 87.36 & 87.08 & 88.80 & \textbf{0.50} & \textbf{0.7} & 74.02 & 74.37 & 97.0 & \textbf{0.69} & 89.03 & 88.95 & 2.25 & 1.50 \\
        & DROP (ours) & 77.06 & 76.66 & 75.03 & \textbf{1.20} & 1.60 & 1.1 & 65.78 & 65.48 & \textbf{1.20} & 1.02 & 89.36 & 89.45 & \textbf{0.75} & \textbf{1.0} \\
        \hline
        \multirow{4}{*}{Chameleon} & Undefended & 87.06 & 87.34 & 87.43 & 100.0 & 100.0 & 100.0 & 74.76 & 74.34 & 99.96 & 99.91 & 89.02 & 88.98 & 99.25 & 99.5 \\
        & FLIP & 83.29 & 83.41 & 83.64 & 96.8 & 97.1 & 98.0 & 70.31 & 70.01 & 88.91 & 97.97 & 88.48 & 88.65 & 77.25 & 25.25 \\
        & DROPlet (ours) & 87.72 & 87.1 & 87.33 & 99.8 & 99.9 & \textbf{0.4} & 74.11 & 74.4 & \textbf{0.79} & \textbf{0.72} & 88.98 & 88.95 & 1.25 & 1.25 \\
        & DROP (ours) & 75.71 & 75.37 & 76.14 & \textbf{1.0} & \textbf{1.2} & 1.8 & 65.98 & 66.06 & 1.48 & 1.87 & 89.32 & 89.43 & \textbf{1.0} & \textbf{1.0} \\
        \bottomrule
        \end{tabular}
    }
    \caption{\revision{MTA (\%) and ASR (\%) metrics for different attack strategies and at different Data Poisoning Rates (DPR) for the CIFAR-10, \revision{CINIC-10}, and EMNIST datasets. The MCR is fixed at 20\%. DROP achieves low ASR for all DPR values across all attacks. DROPlet maintains higher MTA than DROP and remains robust in most cases, though it is occasionally affected by Chameleon’s adaptive poisoning (e.g., 2.5\% DPR, CIFAR-10). The learning configurations are C4 for CIFAR-10 (same for CINIC-10) and C5 for EMNIST. Refer to \cref{tab:fl_setup_exps} and \cref{tab:fl_setup_exps_emnist} for learning configuration details.}}
    \label{tab:fl_diff_attacks}
\end{table*}

\shortsection{Attacks} We evaluate our defense against a targeted backdoor attack (described in \Cref{sec:threat_model}), where adversarial clients poison inputs from a specific victim class \(y_{\text{victim}}\) to embed malicious behavior in the global model. We evaluate over BadNets \citep{gu_badnets}, \revision{as well as two state-of-the-art adaptive variants: Neurotoxin \cite{neurotoxin} and Chameleon \cite{chameleon}. Neurotoxin improves durability by constraining poisoned updates to model parameters that typically change little during training, embedding a backdoor that persists even after the adversary drops out. Chameleon uses contrastive learning to adapt poisoned inputs to the feature space of benign peer samples, aligning with facilitator images and evading detection. Both attacks are stealthy---causing minimal degradation in clean accuracy---and durable, maintaining high attack success rates across many training rounds.} The adversary applies a trigger \(\mathbf{t}\) to a fraction of victim-class inputs and flips their labels to a target class \(y_{\text{target}}\). Attack success is measured by the \textit{Attack Success Rate (ASR)}, while maintaining high \textit{Main Task Accuracy (MTA)} ensures stealthiness. We evaluate robustness by varying the \textbf{DPR} at 1.25\%, 2.5\%, and 5\%, and the \textbf{MCR} at 20\% and 40\%, to simulate a range of adversarial compromise.

\shortsection{Baselines} To benchmark the effectiveness of DROP, we compare its performance against representative FL poisoning defense mechanisms from the literature. Median Aggregation \cite{yin2018byzantine} is a classical robust aggregation technique that uses the coordinate-wise median of client updates to neutralize the effect of outliers. Multi-Krum \cite{blanchard2017machine} selects and aggregates the most ``trustworthy" client updates by iteratively excluding updates that are furthest from the majority, making it resilient to Byzantine failures. FLTrust \cite{cao2021fltrust} introduces a server-side reference model to measure the ``trustworthiness" of client updates, ensuring that only updates aligning with the reference model are included in the aggregation. FoolsGold \cite{fung2018mitigating} employs similarity-based clustering to mitigate the impact of poisoned updates by reducing their contribution in the aggregation process. FLAME \cite{nguyen2022flame} is a robust aggregation method designed to address both Byzantine and poisoning attacks in FL by selectively penalizing anomalous updates. FLIP \cite{zhang2023flip} estimates client-side trigger reconstruction using adversarial training combined with low-confidence refusals. Finally, FLARE \cite{wang2022flare} leverages penultimate layer representations (PLR) to evaluate the trustworthiness of model updates, assigning trust scores based on PLR discrepancies and aggregating updates weighted by these scores to defend against model poisoning attacks.
For further implementation details on the baselines, please see \Cref{app:baseline_details}.

Our evaluation reveals the strengths and weaknesses of existing defenses and demonstrates the superior robustness of our proposed defense, DROP, across various attack scenarios.

\subsection{Robustness Across FL Learning Configurations}

A key strength of DROP is its robustness across diverse FL learning configurations, ensuring reliable protection against targeted backdoor attacks independent of the specific learning parameter settings. While other defenses exhibit significant performance variability depending on the enforced learning parameters (\Cref{fig:baselines_heatmap}), DROP consistently maintains near-zero ASR across all configurations of interest, as shown in \Cref{tab:drop_across_configs}. This consistency demonstrates DROP’s adaptability and makes it a learning configuration agnostic defense, capable of handling varying client learning rates, batch sizes, and local training epochs without compromising on security. This robustness is particularly important for real-world FL deployments, where learning configurations can vary dynamically across tasks. By ensuring reliable performance across a wide range of setups, DROP offers practical, consistent protection against adversarial threats, making it well-suited for heterogeneous FL environments.

\begin{table}[ht]
    \small
    \centering
    \resizebox{\columnwidth}{!}{
        \begin{tabular}{lcc|lcc}
        \toprule
        \textbf{Config} & \textbf{MTA (\%)} & \textbf{ASR (\%)} & \textbf{Config} & \textbf{MTA (\%)} & \textbf{ASR (\%)} \\
        \midrule
        C1 & 69.47 & 1.8 & C6 & 62.61 & 1.6\\
        C2 & 71.77 & 0.7 & C7 & 66.17 & 1.7\\
        C3 & 80.20 & 2.8 & C8 & 64.71 & 1.6\\ 
        C4 & 76.05 & 1.3 & C9 & 80.44 & 0.8\\
        C5 & 75.24 & 1.5 & C10 & 80.50 & 5.5\\
        \bottomrule
        \end{tabular}
    }
    \caption{MTA (\%) and ASR (\%) for DROP on CIFAR-10 for 1.25\% DPR and 20\% MCR, across all 10 configurations identified in \Cref{sec:defenses_fail_across_configs}. DROP achieves low ASR {\bf across all configurations}.}
    \label{tab:drop_across_configs}
\end{table}

\begin{figure*}[ht]
    \centering
    \includegraphics[width=0.97\linewidth]{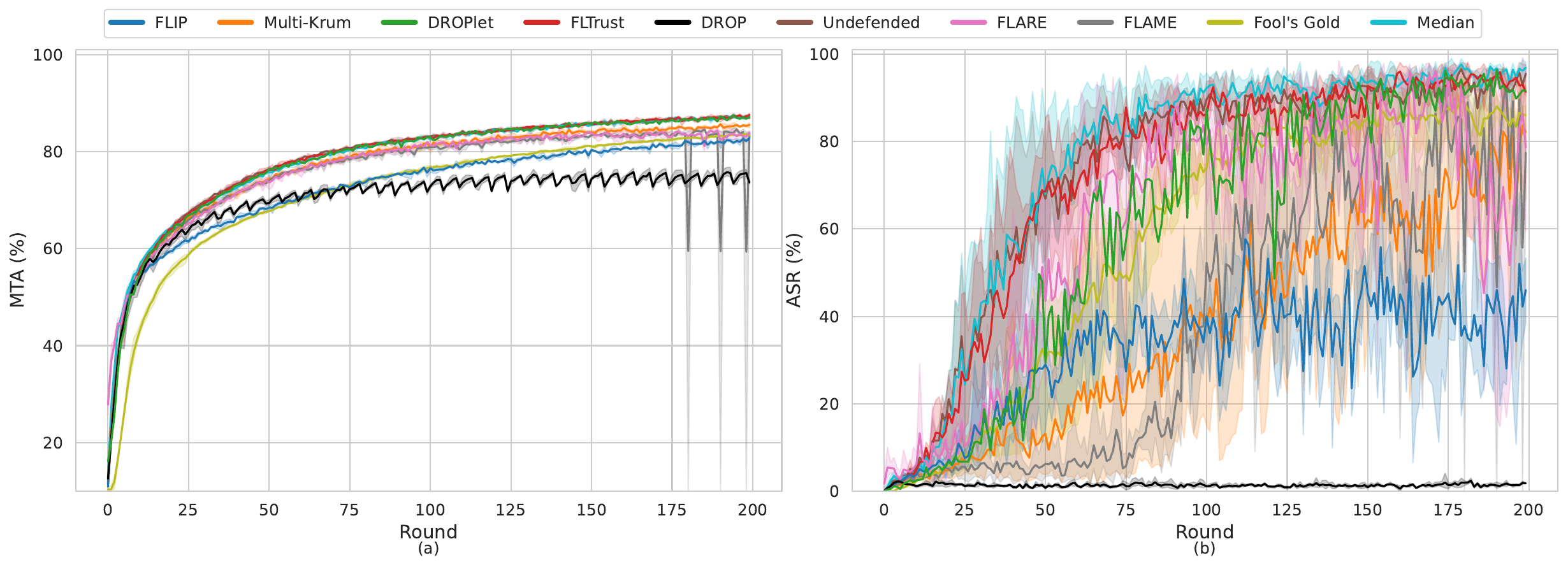}
    \caption{MTA (a) and ASR (b) across rounds for various defenses, for CIFAR-10 with 1.25\% DPR and 20\% MCR for configuration C4. Certain defenses like FLIP and FLAME have wildly fluctuating ASR across rounds, making them unreliable. DROP instead achieves consistently low ASR in all rounds.}
    \label{fig:multiple_round_progression}
\end{figure*}

\shortsection{Consistency Across Rounds} A robust defense must ensure low ASR during rounds where the MTA remains acceptable, preventing randomness in ASR fluctuations from determining the outcome of FL training. While some baselines (\eg FLIP, FLAME) initially achieve low ASR, their performance deteriorates over subsequent rounds, resulting in inconsistent suppression of the attack. This variability is problematic because in a real-world setting, FL training might terminate at any round, and reliance on transient phases of low ASR does not guarantee reliable protection. In \Cref{fig:multiple_round_progression}, we compare the performance of various defensive baselines with DROP over 200 rounds. Although some baselines degrade more quickly (\eg Median, FLTrust), others (\eg FLIP, FLAME) exhibit slower degradation, yet suffer from significant ASR fluctuations across rounds. This inconsistency makes it difficult to guarantee robust defense throughout the training process. In contrast, DROP consistently maintains near-zero ASR across all rounds, offering reliable protection regardless of when training might terminate. As described in the Property 2 (\Cref{sec:threat_model}), \textbf{\textit{this stability ensures that practitioners do not have to depend on chance for effective defense, making DROP a practical and dependable solution for real-world FL systems.}}

\subsection{Adaptability to Varying Attack Stealthiness}
Varying DPR simulates different levels of attack stealthiness, while varying MCR represents scenarios where the fraction of malicious clients in the federation changes.

\shortsection{Varying DPR}  
By varying the attacker's DPR, we simulate increasingly stealthy targeted backdoor attacks, with DPR as low as 1.25\%. We observe that all baseline defenses fail to defend against targeted backdoors across CIFAR-10, \revision{CINIC-10}, and EMNIST, regardless of attack stealthiness (\Cref{tab:fl_vary_dpr}).  
Some defenses perform adequately only when the attack is overt. For instance, for CIFAR-10, FLARE mitigates the attack effectively at 5\% DPR, achieving 3.5\% ASR, but fails entirely when the DPR is reduced (\eg 2.5\%). \revision{Similarly, for CINIC-10, FLIP demonstrates moderate robustness at 2.5\% DPR, achieving 38\% ASR, but succumbs to the attack at 1.25\% DPR}. In contrast, DROP consistently defends against targeted backdoor attacks across all levels of DPR, maintaining near-zero ASR (1.5\%, 1.2\%, and 0.5\%) in all configurations.

\begin{table*}[h]
    \centering
    \resizebox{\textwidth}{!}{
        \begin{tabular}{l|c|cc|cc||c|cc|cc||c|cc|cc}
        \toprule
        \multirow{2}{*}{\textbf{Defense}} & \multicolumn{5}{c}{CIFAR-10} & \multicolumn{5}{c}{CINIC-10} & \multicolumn{5}{c}{EMNIST} \\
         & \multicolumn{3}{c}{MTA (\%)} & \multicolumn{2}{c}{ASR (\%)} & \multicolumn{3}{c}{MTA (\%)} & \multicolumn{2}{c}{ASR (\%)} & \multicolumn{3}{c}{MTA (\%)} & \multicolumn{2}{c}{ASR (\%)} \\
         \cline{2-16}
         \multicolumn{1}{r|}{MCR (\%)} & NA & 20 & 40 & 20 & 40 & NA & 20 & 40 & 20 & 40 & NA & 20 & 40 & 20 & 40 \\
        \midrule
        Undefended & 86.48 & 86.56 & 86.05 & 92.1 & 99.6 & 74.71 & 74.49 & 74.47 & 98.65 & 99.6 & 87.74 & 87.78 & 87.65 & 94.75 & 100.0 \\
        \hline
        FLAME & 80.66 & 80.78 & 81.11 & 91.3 & 99.8 & 72.93 & 72.77 & 72.62 & 98.6 & 99.32 & 86.28 & 86.43 & 86.71 & 2.50 & 89.50 \\
        FLARE & 83.63 & 83.28 & 83.74 & 58.3 & 98.9 & 71.07 & 71.09 & 71.65 & 93.02 & 97.27 & 87.05 & 86.72 & 86.52 & 27.50 & 96.50 \\
        FLIP & 80.09 & 82.25 & 83.43 & 11.2 & 72.3 & 67.73 & 69.6 & 71.65 & 38.82 & 92.94 & 87.24 & 87.48 & 87.54 & 12.25 & 84.75 \\
        FLTrust & 86.85 & 87.38 & 87.24 & 99.5 & 99.4 & 74.46 & 74.54 & 74.65 & 98.72 & 99.73 & 88.59 & 88.47 & 88.68 & 78.75 & 99.50 \\
        Fool's Gold & 83.93 & 83.80 & 83.73 & 86.2 & 99.7 & 70.25 & 69.96 & 69.93 & 95.29 & 98.77 & 85.96 & 85.91 & 85.82 & 25.50 & 96.75 \\
        Median & 86.41 & 86.10 & 85.54 & 98.9 & 99.1 & 73.29 & 73.38 & 73.26 & 98.87 & 99.69 & 87.66 & 87.66 & 87.71 & 94.25 & 99.75 \\
        Multi-KRUM & 84.28 & 84.41 & 84.04 & 99.2 & 99.8 & 73.61 & 73.58 & 73.58 & 99.74 & 99.55 & 87.37 & 87.13 & 87.20 & 1.00 & 99.25 \\
        \hline
        DROPlet (ours) & 85.69 & 85.49 & 84.87 & \textbf{0.9} & \textbf{0.8} & 74.62 & 74.27 & 74.65 & \textbf{0.48} & 99.73 & 87.77 & 87.66 & 87.37 & 1.50 & 0.75 \\
        DROP (ours) & 69.7 & 69.11 & 70.71 & 1.3 & 1.4 & 65.4 & 65.72 & 64.97 & 1.22 & \textbf{2.36} & 88.77 & 88.67 & 88.43 & \textbf{0.75} & \textbf{0.75} \\
        \bottomrule
        \end{tabular}
    }
    \caption{MTA (\%) and ASR (\%) metrics under no attack (NA) and at different Malicious Client Ratios (MCR) for CIFAR-10, CINIC-10 and EMNIST datasets. The DPR is fixed at 2.5\%. Existing defenses fail to prevent the attack, particularly at 40\% MCR. DROP is resilient in all cases, despite some reduction in MTA. DROPlet performs adequately in most cases independently of the MCR. The learning configurations are C1 for CIFAR-10 (C4 for CINIC-10) and C9 for EMNIST. Refer to \cref{tab:fl_setup_exps} and \cref{tab:fl_setup_exps_emnist} for learning configuration details.}
    \label{tab:fl_vary_mcr}
\end{table*}

\shortsection{Varying MCR}  
By varying the MCR, we simulate scenarios where the attacker controls a small (20\%) or large (40\%) fraction of clients. A high MCR like 40\% significantly increases the likelihood that malicious clients will outnumber benign ones in certain FL rounds due to the random client selection process. This imbalance enables adversarial updates to dominate the aggregation process, amplifying the attack's effectiveness and exposing vulnerabilities in clustering-based defenses. Once again, we observe that most defensive baselines fail to provide adequate protection against the attack under such conditions (\Cref{tab:fl_vary_mcr}). For instance, FLIP reduces the ASR to 11.2\% and 38.82\% for 20\% MCR on CIFAR-10 \revision{and CINIC-10}, but fails to mitigate the attack at 40\% MCR. FLAME, due to its overly aggressive adaptive clipping scheme, struggles to achieve convergence. Both FLAME and Multi-Krum achieve resilience for 20\% MCR on EMNIST, but do not thwart the attack at 40\% MCR. In contrast, DROP consistently demonstrates resilience across both 20\% and 40\% MCR setups, achieving near-zero ASR while, highlighting its robustness even in highly adversarial settings.

\shortsection{Varying Attacks} \revision{DROP is resilient to stealthy state-of-the-art backdoor attack strategies like Neurotoxin and Chameleon. As shown in \Cref{tab:fl_diff_attacks}, it consistently maintains near-zero ASR (1.87\% or lower) across all datasets and poisoning rates, even when facing these durable and adaptive attacks. This highlights DROP’s ability to eliminate embedded backdoors that persist in the model long after the attacker has left, outperforming other defenses that fail to mitigate these advanced threats.}

\subsection{Robustness - Accuracy Tradeoff} While DROP demonstrates strong effectiveness against backdoor attacks, it comes with a notable trade-off in MTA compared to some baselines. For example, in \Cref{tab:fl_vary_dpr} with a DPR of 1.25\%, DROP achieves an MTA of 74.74\%, which is lower than FLTrust (87.93\%) and Median (87.14\%). Similarly, in \Cref{tab:fl_vary_mcr} with an MCR of 20\%, DROP records an MTA of 68.74\%, compared to 86.88\% for FLTrust. This reduction in accuracy primarily stems from the strict distillation-based mechanism, which aims to ensure suppression of adversarial influences. Notably, unlike clustering or aggregation-only defenses, DROP uses model stealing (i.e., MAZE) to apply knowledge distillation, which contributes significantly to the observed MTA drop while exhibiting strong robustness. The trade-off between robustness and accuracy is a well-documented challenge in adversarial machine learning \cite{tsipras2019robustnessoddsaccuracy}.

It is worth noting that one could focus on configurations such as C3, C9, or C10, where we achieve $\sim80\%$ MTA and claim a smaller drop in MTA. While it would indeed be preferable for a practical FL system to utilize such learning configurations, the ability to select configurations without sacrificing protection is a unique strength of our defense. By prioritizing security, DROP is a more reliable choice for high-stakes applications where even a minor vulnerability could lead to significant consequences.

It is also worth highlighting that advancements in knowledge distillation techniques and model-stealing attacks could directly enhance DROP. Improved versions of these methods can serve as drop-in replacements for the current MAZE component, potentially reducing the accuracy trade-off while preserving DROP’s robust defensive capabilities.

\shortsection{DROPlet - Lightweight Variant of DROP}
DROPlet (\Cref{subsec:droplet}) is designed for environments where computational efficiency is a priority. Unlike DROP, DROPlet omits the knowledge distillation component, thereby eliminating the associated computational overhead. This makes it particularly suitable for FL deployments with limited resources or scenarios where minimizing server-side processing time is critical. Notably, execution time for a single non-IID run of DROPlet is comparable to other baselines, completing in approximately 30 minutes, whereas DROP takes 2 hours. \revision{This minimal increase in runtime comes with the added benefit of robust protection against attacks, whereas some defenses like FLIP take up to 57 hours for similar setups}.

While DROPlet may not perform as well as DROP against \textit{certain} highly stealthy, low-DPR attacks (e.g., CIFAR-10, 1.25\% DPR in \Cref{tab:fl_vary_dpr}) \revision{or high-MCR attacks (e.g., CINIC-10, 40\% MCR in \cref{tab:fl_vary_mcr})}, it still provides strong protection in many practical configurations, achieving near-zero ASR in less covert attack scenarios and maintaining robustness across a broad range of learning setups (\Cref{tab:fl_vary_dpr,tab:fl_vary_mcr}).

\revision{While DROPlet provides strong protection in many scenarios, we acknowledge that an effective defense should remain robust regardless of the adversarial strategy. This limitation becomes especially evident in extremely stealthy attacks such as Chameleon (e.g., CIFAR-10, 2.5\% DPR in Table~\ref{tab:fl_diff_attacks}), where DROPlet’s effectiveness drops significantly.} %
Nevertheless, its lightweight design offers significant practical value as a plugin to existing FL defenses, requiring minimal code modifications while enhancing robustness without incurring substantial overhead. By striking a balance between computational efficiency and robustness, DROPlet provides a viable defense solution for resource-constrained FL settings.

\begin{table*}[th]
    \centering
    \resizebox{\textwidth}{!}{
    \begin{tabular}{l|cccc|cccc|cccc|cccc}
    \toprule
    \multirow{2}{*}{\textbf{Defense}} & \multicolumn{8}{c}{CIFAR-10} & \multicolumn{8}{c}{EMNIST} \\
     & \multicolumn{4}{c}{MTA (\%)} & \multicolumn{4}{c}{ASR (\%)} & \multicolumn{4}{c}{MTA (\%)} & \multicolumn{4}{c}{ASR (\%)} \\
     \cline{2-17}
     \multicolumn{1}{r|}{$Dir(\alpha)$} & 1 & 10 & $10^2$ & $\infty$ & 1 & 10 & $10^2$ & $\infty$ & 1 & 10 & $10^2$ & $\infty$ & 1 & 10 & $10^2$ & $\infty$ \\
    \midrule
    Undefended & 86.30 & 87.32 & 87.07 & 87.55 & 94.7 & 86.3 & 95.3 & 85.5 & 87.29 & 87.72 & 87.53 & 87.78 & 93.00 & 94.50 & 93.50 & 94.75  \\
    \hline
    FLAME & 59.43 & \textcolor{red}{10.00} & 82.34 & 84.86 & \textbf{0.4} & \textcolor{red}{0.0} & 31.5 & 88.7 & 85.91 & 86.48 & 85.54 & 85.86 & \textbf{13.5} & 56.75 & 3.0 & 3.25  \\
    FLARE & 83.81 & 81.37 & 82.55 & 85.06 & 48.3 & 59.9 & \textbf{15.4} & 94.0 & 86.48 & 86.24 & 86.57 & 86.86 & 84.0 & 28.25 & 3.75 & 63.5 \\
    FLIP & 78.50 & 81.90 & 83.1 & 83.09 & 54.0 & \textbf{25.4} & 47.8 & 53.4 & 87.36 & N/A & N/A & 88.52 & 14.75 & N/A & N/A & 46.50 \\
    FLTrust & 86.40 & 86.84 & 85.34 & 87.93 & 99.2 & 97.4 & 76.6 & 90.4 & 88.21 & 88.43 & 88.36 & 89.34 & 92.5 & 94.5 & 95.75 & 97.75 \\
    Fool's Gold & 80.00 & 82.93 & 83.95 & 83.85 & 79.7 & 73.1 & 87.0 & 83.0 & 85.51 & 85.76 & 85.70 & 86.71 & 28.25 & 20.75 & 17.50 & 41.75 \\
    Median & 86.32 & 86.31 & 87.10 & 87.48 & 97.3 & 93.5 & 98.3 & 95.3 & 87.24 & 87.61 & 87.51 & 87.72 & 92.0 & 91.5 & 91.5 & 97.0 \\
    Multi-KRUM & 82.74 & 84.56 & 85.15 & 85.96 & 97.8 & 98.2 & 73.9 & 86.0 & 86.49 & 86.75 & 86.84 & 87.11 & 80.75 & 16.75 & 2.5 & 1.0 \\
    \hline
    DROPlet (ours) & 86.62 & 86.08 & 86.99 & 87.49 & 94.5 & 94.4 & 96.1 & 91.3 & 87.14 & 87.51 & 87.74 & 87.66 & 93.0 & 91.75 & 86.0 & 1.5 \\
    DROP (ours) & 69.28 & 73.05 & 72.42 & 76.05 & 88.7 & 93.0 & 80.6 & \textbf{1.3} & 87.66 & 84.48 & 84.16 & 88.57 & 16.50 & \textbf{4.75} & \textbf{1.75} & \textbf{0.75} \\
    \bottomrule
    \end{tabular}
    }
    \caption{MTA and ASR across rounds for various defenses for non-IID data with varying $Dir(\alpha)$, for 1.25\% DPR and 20\% MCR, using configuration C4 for CIFAR-10  and C9 for EMNIST. FLAME fluctuates wildly and often ends up with models that have not learned anything effectively, leading to random-guess MTA (\textcolor{red}{red}) and consequently, zero ASR. Results for FLIP are shown only for $Dir(\alpha) = 1$ due to its extreme computational overhead, as the method requires adversarial training for each client at every round. As observed, DROP is the most resilient defense on EMNIST and all  defenses  struggle  on the more complex CIFAR-10 learning task.
    }
    \label{tab:noniid}
\end{table*}

\subsection{Hardness Under Non-IID Settings}
\label{sec:non_iid}
The non-IID nature of client datasets introduces significant challenges for defending against targeted backdoor attacks, as data distributions among clients can be highly imbalanced. In such settings, the imbalance and diversity of client data make it difficult to distinguish whether differences in model updates arise from the non-IID nature of the data or from benign versus malicious clients. This makes it difficult for clustering-based defenses to differentiate between malicious and benign updates, thereby increasing the stealthiness of the attack. Consequently, defenses that perform well under IID conditions often exhibit reduced effectiveness when faced with non-IID data, underscoring the need for adaptive approaches capable of handling diverse and heterogeneous client data.

\Cref{tab:noniid} presents the results of evaluating various defenses under increasingly non-IID conditions for CIFAR-10 and EMNIST. While both datasets are commonly used in FL research, CIFAR-10 poses a greater challenge due to its higher intra-class diversity, which allows malicious updates to blend more seamlessly with legitimate ones, making it harder to detect and defend against backdoor attacks. This complexity is reflected in the relatively poor performance of most defenses on CIFAR-10 under non-IID conditions. FLIP for instance, while reducing ASR compared to most other defenses, still suffers from significant fluctuations in ASR, sometimes exceeding 60\% under non-IID conditions. This variability highlights the challenge of achieving reliable defense in high-stakes applications where consistency is crucial. 

The results for DROP, however, reveal a nuanced performance. For EMNIST, DROP maintains near-zero ASR across a range of $\alpha$ values, with a slight increase to 16.5\% at the most skewed setting ($\alpha=1$). This demonstrates DROP’s strong resilience in handling less complex, smaller-scale datasets. On the other hand, for CIFAR-10, DROP struggles, with ASR fluctuations persisting even under relatively moderate non-IID conditions. We believe this performance gap may stem from the inherent difficulty of the CIFAR-10 task, which, due to its high class diversity and the complex nature of the data, challenges any defense’s ability to effectively distinguish between benign and adversarial updates.

While our defense performs well in many non-IID scenarios, the fluctuating ASR observed in more complex tasks like CIFAR-10 underscores the \revision{general} difficulty of defending against backdoor attacks under such conditions. These results suggest that \textbf{more research is needed to develop defenses capable of better identifying malicious updates in the presence of significant data heterogeneity}.

\section{Conclusion}
\label{sec:conclusion}

Our study highlights the critical role that learning configurations and attacker strategies play in the success of targeted backdoor attacks in FL. Existing defenses often focus on narrow and fixed configurations, making it difficult to assess their robustness under realistic and diverse setups while the performance across different learning configurations (\ie local learning rate, batch size, training epochs) varies dramatically. We believe future evaluations should consider a broader range of configurations to better capture real-world deployment scenarios. Our results provide a starting point by identifying "danger zones"—configurations that enable stealthy attacks to succeed—and we encourage researchers to test their defenses across these configurations to better understand where defenses are effective and where they fail. This approach ensures that robustness claims are not limited to specific configurations but hold across diverse settings. We recommend future evaluations to include, among other configurations, low learning rates and small batch sizes, as these have been shown to increase the stealthiness of backdoor attacks.

To address this gap, we propose DROP, a novel defense mechanism that is agnostic to learning configurations and the adversary's stealthiness. By using this novel architecture based on clustering, activity tracking, and knowledge distillation, DROP offers a comprehensive solution for mitigating targeted backdoor attacks, ensuring reliable protection across diverse FL setups. Our extensive evaluation across multiple datasets, varying data distributions, \revision{attacks}, and a wide range of attacker configurations highlights DROP's robustness. Unlike existing baselines, which often fail under certain learning configurations or in the presence of stealthy attacks, DROP consistently achieves near-zero attack success rate.

By adopting an exhaustive evaluation methodology and releasing a comprehensive codebase%
, we hope to provide a testbed for future researchers to rigorously assess new defenses and foster advancements in the field. Our work lays the groundwork for designing robust, universal defense mechanisms that can generalize across diverse FL environments, contributing to the long-term security and reliability of federated learning systems.

\ifpreprintversion
\section*{Acknowledgments}
This research was supported by the Department of Defense
Multidisciplinary Research Program of the University Research
Initiative (MURI) under contract W911NF-21-1-0322, and by NSF under grants CNS-2312875 and CNS-2331081.

\fi

\bibliographystyle{IEEEtran}
\bibliography{main}

\appendix
\subsection{Bound Analysis on Number of Malicious Clients per Round}
\label{app:bound_analysis}

Let \( N \) denote the total number of clients and \( M \) the number of malicious clients. The ratio of malicious clients is given by the Malicious Client Ratio (MCR), \( \rho = \frac{M}{N} \). In each round of Federated Learning (FL), we sample \( C \) clients randomly from the \( N \) total clients.
Let $\mathcal{M}$ be the number of malicious clients (out of $C$) sampled in some round. We are interested in finding a lower bound on the probability that there are more malicious clients than benign ones in a round of FL, \ie \( P\left(\mathcal{M} \geq \frac{C}{2}\right) \).

The selection of malicious clients can be modeled as a Binomial distribution where probability of success is the MCR ($\rho$). Thus,  our desired probability can be written in terms of the CDF $F$ of this distribution:
\begin{align}
    P\left(\mathcal{M} \geq \frac{C}{2}\right) = 1 - F\left(\frac{C}{2}, C, \rho\right).
\end{align}
Using a Chernoff bound \citep{arratia1989tutorial}, we can thus obtain a lower bound on the probability that malicious clients outnumber benign ones in a given round as:
\begin{align}
    & P\left(\mathcal{M} \geq \frac{C}{2}\right) \nonumber \\
    &\geq 1 - \exp\left(-C \cdot \left(\frac{1}{2}\left(\ln\frac{1}{2\rho} + \ln\frac{1}{2(1-\rho)}\right)\right)\right) \nonumber \\
    & = 1 - \exp\left(C \cdot \frac{1}{2} \ln\left(4\rho(1-\rho)\right)\right) \nonumber \\
    & = 1 - \left(4\rho(1-\rho)\right)^{\frac{C}{2}}.
\end{align}

\subsection{Related Work}
\label{sec:related_work}

The increasing adoption of FL has spurred extensive research on defending against poisoning attacks, particularly backdoor attacks, which pose a serious threat due to their ability to embed hidden malicious behaviors without degrading the model’s overall performance. Several defense mechanisms have been proposed to mitigate these threats, including robust aggregation methods, trust-based filtering, and anomaly detection techniques. Despite these efforts, existing defenses often exhibit limitations when evaluated across diverse attack configurations, especially under varying data poisoning rates, malicious client ratios within the federation and non-IID client data distributions.

\textit{Robust aggregation-based defenses.} Coordinate-wise median aggregation \citep{yin2018byzantine} and Multi-Krum \citep{blanchard2017machine} are two classical approaches designed to mitigate Byzantine failures by robustly aggregating model updates. The coordinate-wise median computes the median value for each model parameter across client updates, neutralizing the impact of outliers. Multi-Krum, on the other hand, iteratively selects and aggregates updates that are closest to the majority based on pairwise distances, making it resilient to adversarial updates. While these methods are effective against simple poisoning attacks, they struggle to defend against stealthy backdoor attacks, particularly when the data poisoning rate is low or the client data distribution is highly skewed. Additionally, both methods assume IID client data, limiting their robustness in realistic non-IID FL settings.

\textit{Trust-based defenses.} FLTrust \citep{cao2021fltrust} introduces a server-side reference model trained on a small trusted dataset to measure the trustworthiness of client updates. Only updates that align closely with the reference model are aggregated, providing a strong baseline against various adversarial strategies. However, FLTrust’s performance can degrade when the trusted dataset is not fully representative of the overall data distribution and if the attack is very stealthy.  

\textit{Similarity and anomaly detection-based defenses.} Fool’s Gold \citep{fung2018mitigating} uses similarity-based clustering to detect and penalize clients that contribute updates with similar gradients across multiple rounds, under the assumption that adversaries tend to behave similarly. While this approach reduces the contribution of malicious clients, its reliance on similarity metrics makes it susceptible to adaptive adversaries that can evade detection by introducing slight randomness in their updates. FLAME \citep{nguyen2022flame}, which employs adaptive clipping to detect and filter anomalous gradients, also shows promise in defending against targeted attacks. However, as we observe in our experiments, FLAME’s adaptive clipping can become overly aggressive, preventing convergence in some scenarios.  

\textit{Penultimate layer representation-based defenses.} FLARE \citep{wang2022flare} leverages discrepancies in penultimate layer representations (PLR) of model updates to assign trust scores and filter out potentially malicious updates. This approach works well for overt poisoning attempts but, as we demonstrate, struggles to mitigate stealthy attacks with low DPR, where poisoned updates closely resemble benign ones.  

\textit{Adversarial training-based and sample rejection defenses.} FLIP \citep{zhang2023flip} represents a different class of defenses by introducing the ability to reject individual samples during aggregation. By leveraging adversarial training and low-confidence refusals, FLIP aims to reconstruct client-side triggers and reject poisoned updates. While effective to some extent, FLIP’s rejection mechanism often discards a non-negligible fraction of clean samples, which can negatively impact the main task accuracy (MTA). As observed in our experiments, this trade-off in rejection-based defenses can lead to deceptively high MTA metrics, as benign samples that are incorrectly rejected are excluded from evaluation. In contrast, our proposed defense does not require such sample-level rejection, thereby preserving the overall integrity of the clean data and achieving a better balance between MTA and ASR across varying configurations. A crucial consideration when designing FL frameworks is that participating nodes often have limited computational resources, which can constrain the feasibility of complex operations. \eg FLIP relies on client-side adversarial training—a computationally expensive process. While this approach is viable in their setup, where only 10 clients are sampled per round, increasing the number of selected clients significantly extends the duration of each round, making it impractical for large-scale deployments.

Despite the progress made by these defenses, a significant gap remains: no existing method consistently performs well across different DPR levels, IID and non-IID data distributions, and varying MCR. Our proposed defense, DROP, addresses these issues by providing a robust, configuration-agnostic solution that is effective across a broad range of attack and learning configurations.

\subsection{Experimental Details}
\label{app:exp_details}

\subsubsection{Baselines}
\label{app:baseline_details}

Each of the defense methods which were presented aims to limit the influence of malicious updates during aggregation. However, most works provide limited or inconsistent details about their evaluation setups, particularly concerning client learning configurations such as learning rate, batch size, and the number of local training epochs. For instance, Median \citep{yin2018byzantine} is evaluated on a simpler learning task (MNIST \cite{mnist}) and specifies only the total number of participating clients, without providing key details about the client learning setup, such as the learning rate, batch size, or the number of local epochs. Similarly, Multi-Krum \citep{blanchard2017machine} reduces the impact of outliers using robust statistics but primarily focuses on how the data is partitioned among clients. While it does evaluate the method across different batch sizes, it lacks a detailed discussion of the broader local training setup. FLTrust \citep{cao2021fltrust} adopts a trusted server-side reference model to filter anomalous updates. The authors report using a "combined" learning rate of 0.002, a batch size of 64, and a single local training epoch. In contrast, FoolsGold \citep{fung2018mitigating}, which identifies and penalizes suspiciously similar client contributions, does not explicitly report the learning rate or the number of local epochs in its evaluation, only mentioning batch sizes of 10 or 50 depending on the dataset. FLAME \citep{nguyen2022flame}, which employs anomaly detection to flag potentially malicious gradients, describes the structure of the federation but omits critical information about the local learning setup, such as learning rate, batch size, or the number of epochs. FLIP \citep{zhang2023flip}, on the other hand, presents a more detailed setup. We adapted the original implementation and hyperparameters for our evaluation. The original defense assumes that both the defense and malicious client poisoning are triggered after model convergence, citing interference in convergence if malicious activity begins earlier. However, our observations show that targeted backdoors do not disrupt model convergence even when initiated at the beginning of FL training. Thus, starting the defense only after convergence is ineffective since poisoning has already occurred by that point. On the other hand, initiating the defense too early leads to suboptimal performance. Therefore, we activate the defense after the first 10 rounds for CIFAR-10 and after the first 3 rounds for EMNIST to balance effectiveness and performance.

\subsubsection{DROP Parameters}
\label{sec:drop_params}

Knowledge distillation, particularly in the context of model stealing attacks, is inherently imperfect and cannot replicate the target model exactly, resulting in a minor decrease in MTA. To address this and ensure convergence in the FL setting, the knowledge distillation component is applied every \(K\) rounds instead of every round, allowing the system to recover lost MTA during intermediate rounds. The budget parameter in model stealing attacks and knowledge distillation determines the number of queries used to generate synthetic samples, which are then employed to guide the distillation process. A sufficient query budget ensures the generation of high-quality synthetic data that aligns closely with the target model’s decision boundaries, thereby enhancing the effectiveness of knowledge distillation. For CIFAR-10, we set \(K = 5\) with a query budget of 5M queries. For EMNIST, \(K = 40\) with a query budget of 4M queries. These values strike a balance between computational efficiency and the quality of the distilled global model.

\subsubsection{EMNIST Grid-Search}

\begin{figure}[ht]
    \includegraphics[width=.98\linewidth]{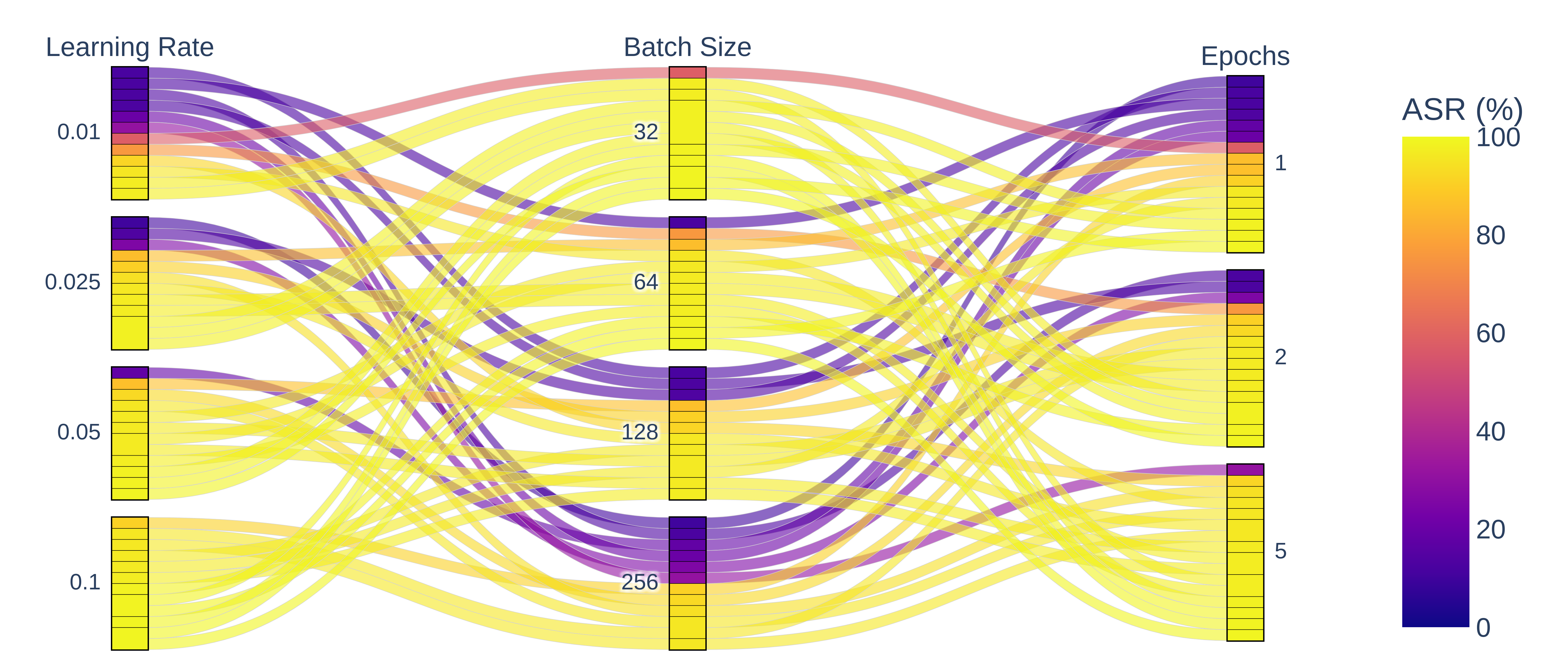}
    \caption{Visualizing the impact of the FL setup (particularly the learning-rate, batch-size, and number of epochs used by clients) on main-task accuracy attack success rate when poisoned clients aim to inject a targeted backdoor for the EMNIST dataset.}
    \label{fig:fl_setup_impact_emnist}
\end{figure}

In the same fashion as \cref{sec:fl_setup_matters} for CIFAR-10, we conduct a grid-search analysis over key hyperparameters, varying the client’s learning rate, batch size, and number of epochs on the EMNIST \citep{cohen2017emnistextensionmnisthandwritten} dataset.
Our findings in \cref{fig:fl_setup_impact_emnist} indicate that targeted backdoor attacks are more likely to succeed across a wider range of learning parameter combinations, with numerous setups yielding an ASR greater than 80\%. In \cref{tab:fl_setup_exps_emnist}, we highlight ten specific learning configurations where the attack achieves high ASR while maintaining a high MTA, underscoring the vulnerability of these setups to adversarial manipulation.

\begin{table}[ht]
    \centering
    \begin{tabular}{llcc|cc}
    \toprule
    \textbf{Config} & \textbf{LR} & \textbf{BS} & \textbf{Epochs} & \textbf{MTA (\%)} & \textbf{ASR (\%)} \\
    \midrule
    C1 & 0.1 & 32 & 2 & 89.23 & 99.00 \\
    C2 & 0.1 & 64 & 5 & 88.22 & 99.00 \\
    C3 & 0.05 & 32 & 5 & 88.48 & 98.75 \\
    C4 & 0.1 & 32 & 1 & 89.59 & 98.75 \\
    C5 & 0.1 & 64 & 1 & 89.20 & 98.50 \\
    C6 & 0.1 & 32 & 5 & 88.73 & 98.50 \\
    C7 & 0.1 & 64 & 2 & 88.87 & 98.25 \\
    C8 & 0.05 & 32 & 1 & 89.21 & 98.25 \\ 
    C9 & 0.01 & 32 & 2 & 87.73 & 97.25 \\
    C10 & 0.025 & 128 & 5 & 86.61 & 96.00 \\
    \bottomrule
    \end{tabular}
    \caption{Client FL configurations for successful stealthy attacks on EMNIST \ie cases where the MTA $\geq 80\%$ and ASR $\geq 95\%$.}
    \label{tab:fl_setup_exps_emnist}
\end{table}
\clearpage

\end{document}